# Exploring the sustainable scaling of AI dilemma: A projective study of corporations' AI environmental impacts

Clément Desroches, Martin Chauvin, Louis Ladan,
Caroline Vateau, Simon Gosset, Philippe Cordier*

*Capgemini Invent*
*145 quai du Président Roosevelt, 92130 Issy Les Moulineaux, France*

## Abstract

The rapid growth of artificial intelligence (AI), particularly Large Language Models (LLMs), has raised concerns regarding its global environmental impact that extends beyond greenhouse gas emissions to include consideration of hardware fabrication and end-of-life processes. The opacity from major providers hinders companies' abilities to evaluate their AI-related environmental impacts and achieve net-zero targets.

In this paper, we propose a methodology to estimate the environmental impact of a company's AI portfolio, providing actionable insights without necessitating extensive AI and Life-Cycle Assessment (LCA) expertise. Results confirm that large generative AI models consume up to 4600x more energy than traditional models. Our modelling approach, which accounts for increased AI usage, hardware computing efficiency, and changes in electricity mix in line with IPCC scenarios, forecasts AI electricity use up to 2030. Under a high adoption scenario, driven by widespread Generative AI and agents adoption associated to increasingly complex models and frameworks, AI electricity use is projected to rise by a factor of 24.4.

Mitigating the environmental impact of Generative AI by 2030 requires coordinated efforts across the AI value chain. Isolated measures in hardware efficiency, model efficiency, or grid improvements alone are insufficient. We advocate for standardized environmental assessment frameworks, greater transparency from the all actors of the value chain and the introduction of a "Return on Environment" metric to align AI development with net-zero goals.

## Introduction

Artificial Intelligence (AI) is transforming industries worldwide, but its recent rapid developments, particularly in Generative AI and Agentic AI, raises urgent sustainability concerns due to their significant environmental impacts[1]. These impacts stem from two primary sources: operational usage, involving electricity and water consumption during training and usage, the manufacturing and the end-of-life processes of hardware equipment. The impacts are linked to the equipment and infrastructures mobilized to deliver Traditional AI and generative AI[2,3] (GenAI) services including servers and IT equipment housed in datacenters, as well as telecommunication networks and end-users devices. Collectively, these contribute to global environmental impacts like greenhouse gas emissions, water consumption, resource depletion (minerals and metals), and the escalating issue and impact of electronic waste.

The energy demands of AI are particularly concerning, with projections from the International Energy Agency (IEA) indicating that global data center electricity consumption could double by 2026, driven by AI and cryptocurrencies[4]. Other studies have examined the projected electricity usage of AI data centers[5–7], all showing potential threefold to eightfold increase. This underscores the urgency of addressing not only energy consumption and associated carbon emissions but also the broader implications of resource usage.

For companies committing to carbon neutrality by 2030 or 2040 in alignment with the Paris Agreement[8], the ability to forecast and address AI's future environmental impacts is critical. In the latest Capgemini Research Institute report study on sustainable GenAI, 64% of companies say AI energy consumption is too complex to measure[1]. Assessing AI's environmental footprint is a challenge exacerbated by its complexity. Operational impacts depend on factors such as model efficiency, carbon intensity of electricity grid that powers data centers, number of users and average usage. Embodied impacts are even harder to evaluate, as they involve intricate complex supply chains, with raw material extraction and transformation, manufacturing of semi-conductors, and eventual hardware disposal.

We usually categorize AI models into two groups: Traditional AI (or task-specific models) and Generative AI (or general-purpose models)[9]. Traditional AI primarily focuses on machine learning (ML) and deep learning models, where users interact with a single model designed for specific predictive tasks (e.g., classification, regression) through input data and output predictions, for computer vision, time series analysis or Natural Language Processing applications. In contrast, a single generative AI model can solve a variety of tasks for users out-of-the-box.[2]

Since 2015, interest in the power consumption and energy efficiency of AI models, particularly deep learning models, has been growing[10–14]. Within the AI system hardware life cycle, the primary focus is on the environmental impacts of manufacturing (i.e., the embodied carbon footprint) and AI use (i.e., the operational carbon footprint). The model development phases include data processing, experimentation, training, and inference. Recent research has substantially studied the environmental impacts of training and inference [9,15–22].

Regarding Generative AI specifically, several studies[9,19,20] directly try and measure emissions using tools like Code Carbon[23] while others, constrained by the lack of transparency from some model providers regarding model architecture, have tried to estimate operational and embodied impacts through simplified models[22,24,25]. Lately, following the recommendation of Strubell et al[20] about the transparency in AI model development phases, the details of training and inference and CO2 emissions of several LLMs are published and closely monitored[18,26,27]. Similarly, several initiatives, such as the LLM-Perf Leaderboard, ML.ENERGY Leaderboard, and Cloud Carbon Footprint, have been established to collect data on the energy consumption of AI models[28–30].

This paper aims to advance the discussion on AI's sustainability by proposing a simplified yet exhaustive methodology to estimate both operational and embodied impacts of AI solutions at a company level. Our methodology comprises four interconnected models:

1. Life cycle impacts of primary components (compute, storage, network) involved in AI projects.

2. Life cycle impacts of AI use cases, categorized for simplicity, focusing on various factors such as energy consumption, GHG emissions, water usage, and resource depletion.

3. AI company portfolio model: we propose a simplified model representing the AI products portfolio of a typical large company portfolio.

4. 2030 AI Landscape projections, that forecasts adoption, efficiency, and complexity of AI technologies up to 2030.

By breaking down these complex assessments into manageable steps, we aim to empower organizations to better understand and project their AI impacts and align their initiatives with global sustainability goals.

# Results

This section presents the results of our modeling and provides an in-depth discussion of the energy consumption of the various AI usage cluster and their multi factor impacts. This includes a comparison of inference and fine-tuning impacts between Generative and Traditional AI use cases, benchmarking against other studies, and an analysis of the relative contributions of embodied versus operational impacts on various factors such as Greenhouse gas (GHG) emissions, water usage or resource depletion. We also share the distribution of impacts of a 2024 fictive company, emphasizing the substantial proportion of generative AI impact within company's operations. Finally, we discuss the 2030 projections of boundaries and intermediate scenarios, illustrating the potential dramatic increase of the impact of Generative AI. We then explore the effectiveness of various mitigation strategies.

## 2024 portfolio impacts

To facilitate the modeling of a typical large company's AI portfolio, we have categorized its use cases into distinct clusters based on five dimensions (type of AI, Use case type, model size, usage frequency and number of users). Details are presented in sections Model 3: Company portfolio model and Company Portfolio Model of supplementary information.

### Use cases' individual impact

Our methodology, based on a dual model (Life cycle impacts of AI hardware components and categorized AI use cases consumption of these primary bricks), accounts for both fine-tuning and inference impacts. In the Table below, we share and compare the inference impacts of these clustered use cases calculated with our methodology (see Methods).

**Table 1 - Electricity consumption (kWh) breakdown (Compute / Storage / network) per inference task by Model Size and Use Case type -** Generative AI models consume significantly more energy per inference compared to Traditional AI: ChatGPT-like applications consume 25 to 4 600 times more energy than conventional NLP use cases. When moving to more companies-specific GenAI applications like RAG or agents, they consume 50 to 25 000 times more energy. While storage and network energy consumption are high for computer vision applications, it remains rather negligible in GenAI tasks. Note that number of output tokens are defined independently from model size, thus storage and network usage remain constant with model size increase. Future research should study more granularly the differences between model sizes.

| Type of AI | Model size | Use Case type | Energy per inference, Compute (kWh) | Energy per inference, Storage (kWh) | Energy per inference, Network (kWh) | Energy per inference, Total (kWh) |
|---|---|---|---|---|---|---|
| **Gen AI** | Low | Chat | 9,30E-05 | 1,68E-08 | 4,59E-08 | 9,31E-05 |
| **Gen AI** | Medium | Chat | 1,55E-03 | 1,68E-08 | 4,59E-08 | 1,55E-03 |
| **Gen AI** | High | Chat | 1,73E-02 | 1,68E-08 | 4,59E-08 | 1,73E-02 |
| **Gen AI** | Low | RAG | 1,54E-04 | 2,86E-07 | 7,79E-07 | 1,56E-04 |
| **Gen AI** | Medium | RAG | 2,64E-03 | 2,86E-07 | 7,79E-07 | 2,64E-03 |
| **Gen AI** | High | RAG | 2,99E-02 | 2,86E-07 | 7,79E-07 | 2,99E-02 |
| **Gen AI** | Low | Agents | 4,97E-04 | 7,99E-08 | 2,18E-07 | 4,97E-04 |
| **Gen AI** | Medium | Agents | 8,54E-03 | 7,99E-08 | 2,18E-07 | 8,54E-03 |
| **Gen AI** | High | Agents | 9,58E-02 | 7,99E-08 | 2,18E-07 | 9,58E-02 |
| **Trad. AI** | *NA* | Tabular | 2,99E-08 | 1,26E-09 | 3,42E-09 | 3,46E-08 |
| **Trad. AI** | *NA* | Computer Vision | 2,58E-05 | 7,81E-05 | 2,13E-04 | 3,17E-04 |
| **Trad. AI** | *NA* | NLP | 3,60E-06 | 2,51E-08 | 6,84E-08 | 3,70E-06 |

### Comparing with other benchmarks and studies

The energy consumption we report for 1 inference of a "small" LLM (0.093 Wh), such as Llama 3.1 8B, aligns closely with empirical energy measurements reported by Luccioni et al.[9] on a similar instance (AWS p4de). Indeed, the energy usage for the facebook/OPT6.7B model 30 reported was 0.082 Wh, while Bloomz-7B[9] consumed 0.104 Wh per inference. Similarly, the reported energy of Bloom 175B in the same study[9] (4 Wh) aligns with the range of medium and large models' energy consumption (Llama 70B: 1.55 Wh and Llama 405B: 17.3 Wh). However, higher energy consumptions are reported by sources like Ecologits[25] and the LLM Perf leaderboard[30]. For instance, the LLM Performance Leaderboard recorded an energy usage of 0.59 Wh for the meta-llama/Meta-Llama-3-8B-Instruct model when generating 208 tokens. This discrepancy could be attributed to differences in generation speed and hardware instances. The LLM Performance Leaderboard reported a generation speed of 23 tok/s on an A10-24GB-150W instance, compared to our study, which achieved 162 tok/s for Azure API.[31]

### Key findings

- Generative AI models consume significantly more energy than traditional AI models, with smaller models like Llama 8B using 25 times more energy than traditional NLP models and larger models like Llama 405B consuming up to 4600 times more.
- The energy demands of larger GenAI models scale substantially**.** A high-sized model, such as Llama 405B, consumes 186 times more energy than Llama 8B for one chat inference. This disparity is primarily attributed to the significantly lower throughput of Llama 405B (down by 83%) and its reliance on a greater number of vGPUs (35x more) related to a 2:1 scaling ratio of memory needed considering FP16. It results in a significant electricity consumption for these large models, each inference reaching *17Wh* (equivalent to toasting bread for 1 minute at 1000 W…).
- Energy consumption also rises with workflow complexity: chat (*1.55 Wh*), Retrieval Augmented Generation (*2.64 Wh*), and agentic workflows (*8.54 Wh*) due to increased computational steps and token generation, highlighting efficiency challenges with larger models and more complex workflows.

- Source of impact differs depending on studied impact and use case (Figure 1) below.

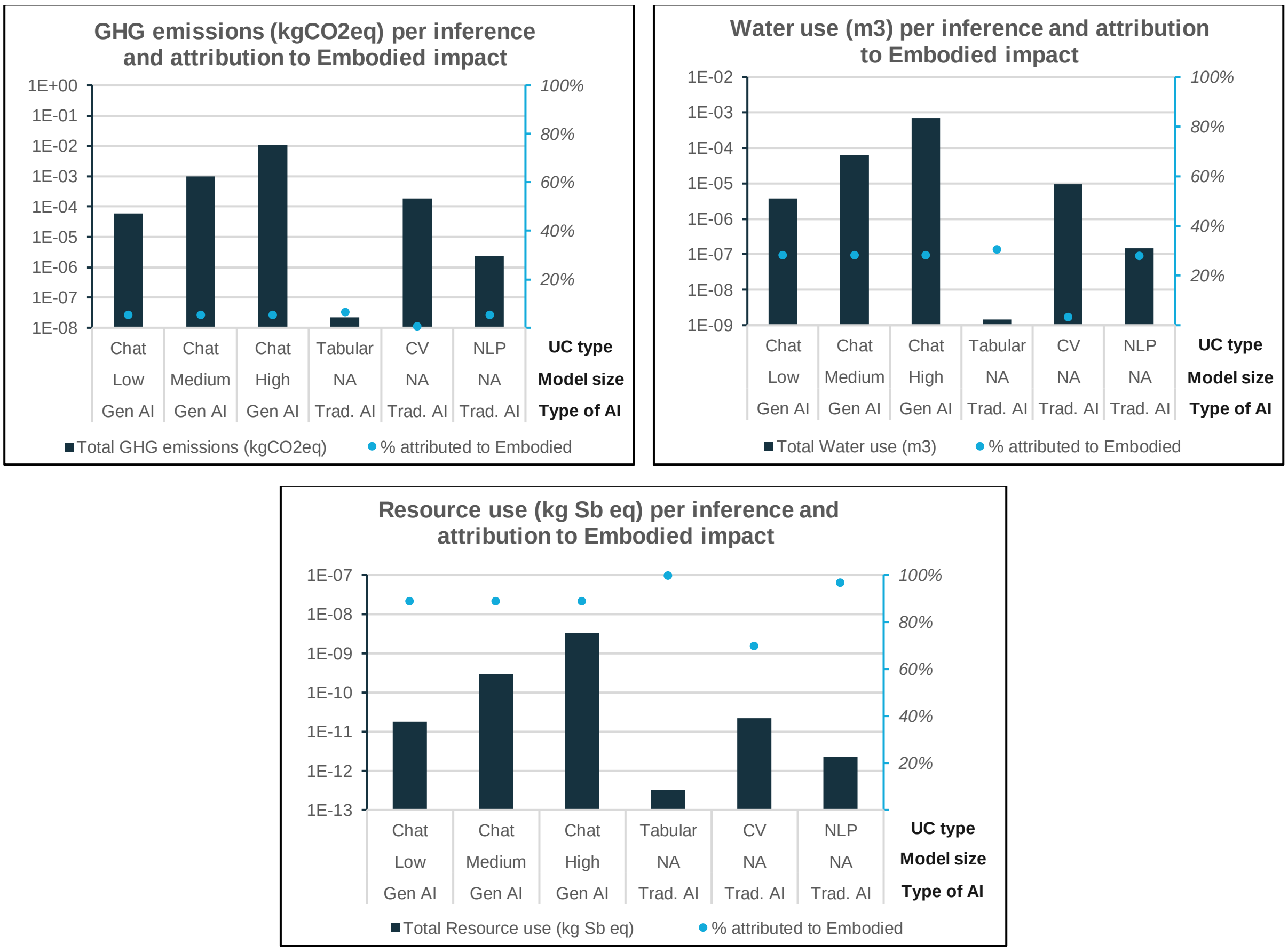

**Figure 1 - GHG emissions, Water use and Resource depletion per inference and attribution to Embodied impact of various use cases -** Water usage is substantially driven by the embodied impacts, particularly the manufacturing of servers. Indeed, the embodied-to-operational ratio (server manufacturing versus electricity usage + datacenter cooling) is much higher on water usage than GHG emissions (28% vs 5% for GenAI). This means that future advancements in compute efficiency to reduce the electricity usage will have a greater impact on GHG reduction than Water usage. Similarly, resource depletion is almost only attributed to the embodied impact reflecting the reliance on rare minerals and resources of servers in their manufacturing process. Note that computer vision applications show a larger operational impact due to the relative importance of storage and network. These variations highlight the differentiated stakes behind every factor. Reducing GHG emissions would require optimizing operational impacts through compute efficiency and electricity grid impact, while resource depletion stakes mainly rely on the efficiency of servers.

## Company Portfolio impact

Across the modeled portfolio, comprising only 29% generative AI (GenAI) and 71% traditional AI use cases, most inferences are attributed to GenAI. This disproportionate contribution arises from the extensive deployment scale (number of users and usage frequency) associated with GenAI applications. Given that GenAI models are inherently more energy-intensive compared to traditional AI models, this trend is further amplified when analyzing energy consumption per use case which scale to 99.9% attributed to generative AI for our representative portfolio.

Regardless of the specific impact metric considered, the inference phase constitutes the most significant contributor compared with model fine-tuning. This predominance is primarily due to the substantial energy demands associated with generative AI (GenAI) use cases. Since most companies are using mainstream LLM like gpt-4o, Claude, Mistral, impact of pre-training[32] phase of Large Language Models (LLMs) has been excluded from this analysis, due to limited transparency on the distribution of general public usages versus corporate-specific applications, it still remains challenging to fairly allocate the environmental impact of pre-training across all users. However, this assumption is not true for LLMs that don't have a substantial usage in production, for which the pre-training phase represent the most important part of its environmental impact. The impact of that kind of models is not discussed here.

The balance between embodied and operational impacts varies significantly depending on the environmental criterion assessed. For example, at the scale of our company portfolio, embodied impacts account for as little as 5% of greenhouse gas (GHG) emissions yet represent as much as 89% of resource depletion and nearly 30% of water usage.

Over a year, the absolute annual impact of the fictive company, considering 100 use cases distributed as explained in the section Company Portfolio Model of supplementary information, stands at an electricity consumption of 3.9GWh for 2,480,000 kgCO2eq. of GHG emissions, 160,000 $m^3$ of Water used and 0.76 kgSbeq.

Assuming all companies listed in the Global 2000 Forbes Index follow a similar use case distribution as proposed, we estimate the combined electricity consumption to be approximately 7.8 TWh in 2024. This figure aligns with the International Energy Agency's (IEA) projections for AI data center electricity[4] estimated to grow from 7.3 TWh in 2023 to 70 TWh by 2026, with the remaining consumption largely driven by public AI usage and other companies.

## Projected 2030 scenario impacts

We have envisaged and analyzed four boundary scenarios and another selected intermediate one to project the evolution of AI's footprint at company level:

1. Scenario ***Steady ascent***: Adoption of generative AI grows gradually with moderate increases in model size and complexity, reflecting conservative market trends. Systemic efficiency remains stable, following historical advancements, leading to a steady rise in energy demand.
2. Scenario ***High Adoption without Boundaries***: Generative AI adoption accelerates significantly, with widespread use of large, complex models and minimal constraints on usage linked to energy consumption and/or pricing. Systemic efficiency sees limited progress, causing substantial energy consumption and environmental impact.
3. Scenario ***Limited Growth with Efficiency Breakthrough*****:** AI adoption remains controlled with modest expansion in use cases and model sizes, emphasizing efficiency over rapid growth. Systemic efficiency improves significantly due to frugal model development, hardware advancements, and adherence to IPCC targets, reducing overall impact.
4. Scenario ***Technological Breakthrough*****:** Generative AI adoption expands rapidly but remains focused on high-sized models and complex usage patterns. Systemic efficiency is driven by groundbreaking advancements in hardware performance, minimizing energy consumption despite increased adoption.
5. ***Intermediate Scenario*****:** AI adoption grows moderately with balanced use case expansion and model complexity, following average market trends. Systemic efficiency sees gradual improvements in hardware performance and policies-aligned electricity impacts, resulting in environmental rising at a moderate pace.

On the usage axis, our findings indicate substantial growth in the number of AI use cases across all scenarios, with increases ranging from a factor of 3.4 to 5.7 (see Table 32 of supplementary information). While generative AI constituted 29%[33] of the company's AI portfolio in 2024, projections suggest that this category will represent half of the portfolios by 2030.

Supported by the significant growth in usage across all scenarios, we find most scenarios display a great increase in environmental footprint, see **Error! Reference source not found.**.

**Table 2 - Environmental footprint of 2030 scenarios indexed on our 2024 portfolio.** Except the Limited growth with efficiency breakthrough, all scenarios, shows a very significant increase of GenAI on all environmental impact, ranging in average from a factor 3 in the technological breakthrough scenario to a factor 21.2 in the high adoption without boundaries.

| 2030 Scenario | Usage | Efficiency | Energy usage | GHG-Emission | Water consumption | Primary energy consumption | Resources depletion |
|---|---|---|---|---|---|---|---|
| *Indexed 2024 portfolio* | | | *100* | *100* | *100* | *100* | *100* |
| **Steady Ascent** | Increase | Low | 552 | 421 | 475 | 419 | 535 |
| **High adoption without boundaries** | Explosion | Low | 2440 | 1862 | 2102 | 1852 | 2363 |
| **Limited growth with efficiency breakthrough** | Increase | High | 30 | 17 | 22 | 17 | 25 |
| **Technological breakthrough** | Explosion | High | 402 | 227 | 288 | 225 | 325 |
| **Intermediate scenario** | High adoption | Low | 755 | 576 | 650 | 573 | 732 |

In the *high adoption without boundaries scenario,* there is a 24.4-fold increase in energy usage and an 18.6-fold increase in GHG emissions, primarily driven by the surge in usage, with a 47% compound annual growth rate (CAGR) for GenAI, 55% CAGR for Agentic AI use cases, and a continuous trend towards larger models with limited improvements in hardware efficiency or electricity mix.

In the *Limited growth with efficiency breakthrough scenario*, which assumes ambitious electricity decarbonization, tremendous advancements in hardware efficiency, including new chip types (such as Neuromorphic chips or Cerebras' Wafer Scale Engines) as well as limited adoption of generative AI and agentic systems (CAGR 32% and 35% respectively), a sustainable reduction of 70% energy usage and 83% GHG emissions is achieved.

Our *Intermediate scenario*, which follows conservative trends in energy efficiency improvements and hardware power density along with median adoption rates (40% GenAI CAGR and 45% Agentic use cases CAGR), shows a substantial increase in both energy usage by a factor of 7.6 and environmental footprint with a 5.8-fold increase in GHG emissions and 6.5-fold increase in water usage.

## Sensitivity analysis

We have performed a sensitivity analysis of main parameters centered around our *Intermediate* scenario to evaluate impact of our model size hypothesis and agentic use cases deployment as well as reflect on the viability of technological only solutions towards sustainable AI usage in line with a 90% GHG reduction compared with 2024.

### Model size

We have varied the parameter describing generative AI model size evolution in 2030 with +/- 10% uncertainty and find our approach translate this uncertainty linearly with a 1:1 factor indicating high dependency on this parameter (Table 3). We expect this factor to increase linearly with a 1:2 factor due to FP16 parameter encoding that requires 2 bytes per parameter, resulting in a doubling of RAM requirements per new parameter and consequently twice the GPU units needed. We have deferred this model improvement for future work.

Based on our findings and the significant impact of this factor, we recommend increased attention be given to this parameter, as limited information is currently available or disclosed by main service providers.

**Table 3 - Sensitivity Analysis on Model Size Projection Parameter (2030 Intermediate Scenario).** This table presents the sensitivity analysis of energy usage under the 2030 Intermediate scenario, focusing on the impact of a 10% uncertainty in the model size projection. The Intermediate scenario assumes that adoption of AI progresses steadily, with a moderate increase in both the range of use cases and model complexity, aligning with typical market trends. Gradual advancements in hardware efficiency and electricity impact management, in line with policy standards, lead to a moderate rise in environmental impact. The current results indicate a linear variation in energy usage. However, a 1:2 variation is expected due to the FP16 parameter. This model refinement has been postponed for future work.

| | Energy usage | GHG Emission | Water consumption | Primary energy consumption | Resources depletion | Projected model size evolution | Energy usage sensitivity |
|---|---|---|---|---|---|---|---|
| *Indexed 2024 portfolio* | *100* | *100* | *100* | *100* | *100* | | |
| **Intermediate scenario** | 755 | 576 | 650 | 573 | 732 | | |
| **Lower projected model size** | 680 | 519 | 585 | 516 | 658 | -10% | -10.0% |
| **Higher projected model size** | 831 | 634 | 716 | 631 | 805 | +10% | +10.0% |

## Agentic adoption

The influence of Agentic AI adoption has been analyzed by examining the impact of the estimated Compound Annual Growth Rate (CAGR) associated with the penetration of associated use cases. Our analysis demonstrates that energy consumption is highly sensitive to the adoption rates of agentic use cases, primarily due to the exponential growth assumed in the CAGR projections (Table 4). We identify a second-order polynomial relationship between CAGR and the energy consumption of AI portfolios, as illustrated in Figure 2.

Considering the expected development of multi-agents use cases, increase model complexity and reasoning models like OpenAI o3, we have also studied the influence of the number of output tokens in our approach (Table 4). Similar to model size, we find that our approach translates the uncertainty in this factor with a 1:1 factor.

**Table 4 - Sensitivity analysis on GenAI agents' adoption and output tokens.** Energy usage sensitivity for the Intermediate scenario after applying uncertainties on its Agentic penetration CAGR and its output size evolution. Depending on penetration rate of Agentic use cases, companies would expect a substantial sensitivity on the consequent electricity usage.

| | Energy usage | GHG emission | Water consumption | Primary energy consumption | Resources depletion | Agentic penetration CAGR | Projected output size evolution | Energy usage sensitivity |
|---|---|---|---|---|---|---|---|---|
| *Indexed 2024 portfolio* | *100* | *100* | *100* | *100* | *100* | | | |
| **Intermediate scenario** | 755 | 576 | 650 | 573 | 732 | 45% | *NA* | 0% |
| **Lower agents adoption** | 512 | 391 | 441 | 389 | 497 | 25% | *NA* | -32.1% |
| **Low agents adoption** | 612 | 467 | 527 | 464 | 593 | 35% | *NA* | -19.0% |
| **High agents adoption** | 958 | 731 | 825 | 727 | 927 | 55% | *NA* | 26.8% |
| **Higher agents adoption** | 1237 | 944 | 1066 | 939 | 1198 | 65% | *NA* | 63.8% |
| **Lower projected number of output tokens** | 680 | 519 | 585 | 516 | 658 | *NA* | -10% | -10.0% |
| **Higher projected number of output tokens** | 831 | 634 | 716 | 631 | 805 | *NA* | +10% | +10.0% |

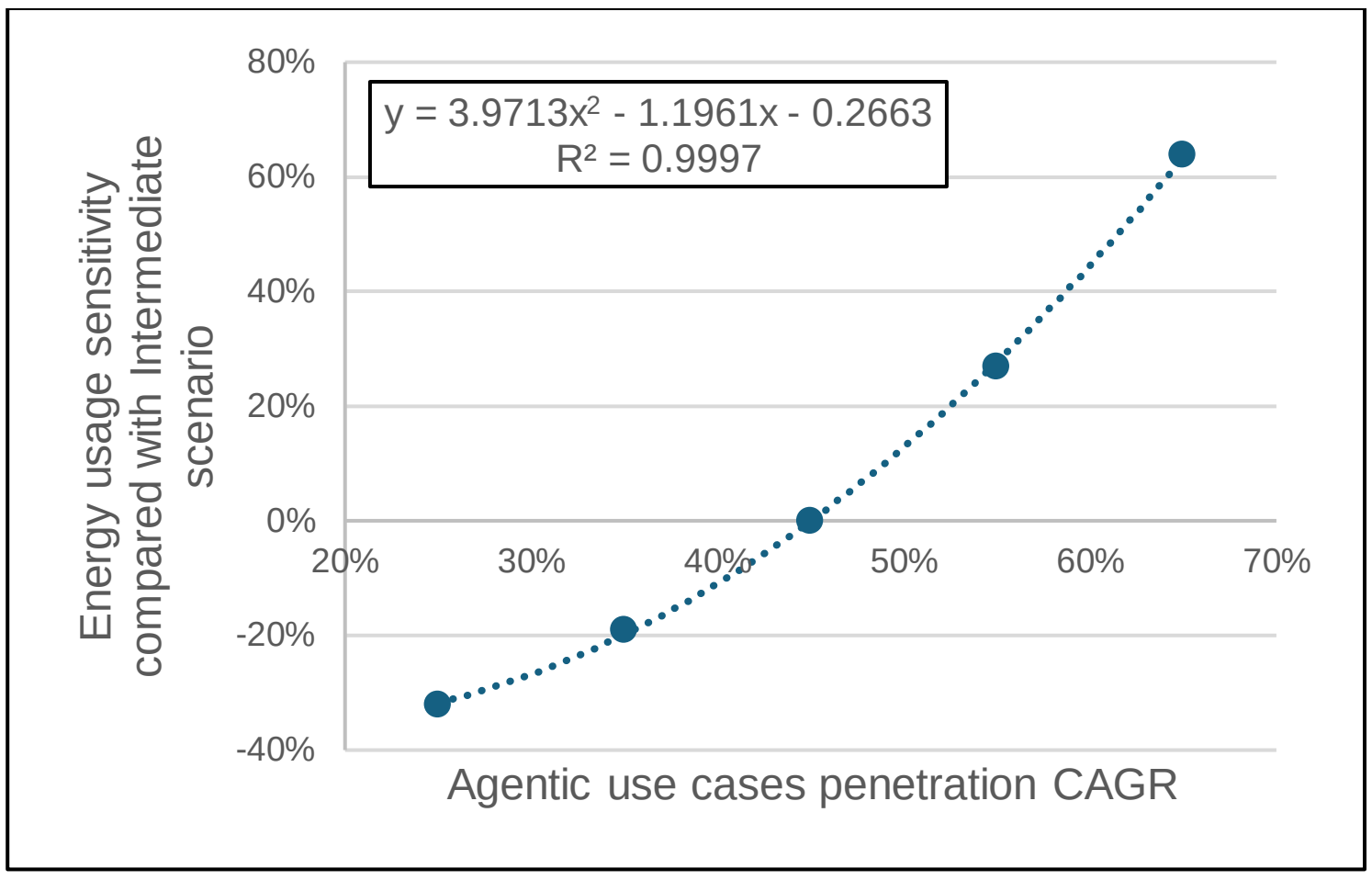

**Figure 2 - Influence of Agentic penetration CAGR on portfolio's energy usage.** The analysis reveals a significant second-order polynomial correlation, indicating that as the Compound Annual Growth Rate (CAGR) of Agentic penetration increases, that, by definition, represents exponential growth over time**,** the energy consumption of AI portfolios exhibits a similar non-linear upward trend, illustrating that the Compound Annual Growth Rate (CAGR), by definition, represents exponential growth over time. This insight emphasizes the importance of managing growth strategies to balance technological advancement with sustainability objectives.

## Hardware efficiency

In this thought experiment, we have explored the feasibility of aligning the *High Adoption Without Boundaries* and *Intermediate* scenarios with 90% GHG reduction compared with 2024, defined as achieving at least a 90% reduction in greenhouse gas (GHG) emissions compared to the 2024 portfolio. This alignment has been attempted through adjustments to key exogenous parameters, specifically Power Usage Effectiveness (PUE), energy mix decarbonization, and hardware efficiency.

Improvement factors for PUE and energy mix were aligned with, respectively, best-in-class current trends in data center efficiency[34] and the International Energy Agency's actual country policy pathways[35] respectively. However, achieving the required reductions also necessitated hardware efficiency improvements by factors of 565x or 175x, depending on the adoption rates of Generative AI (GenAI) and agentic systems (Table 5). These findings underscore the significant challenges associated with meeting a 90% GHG reduction solely through technological advancements.

**Table 5 - Improvement factors on Hardware efficiency to achieve 90% GHG emissions decrease compared with 2024 impact from High adoption without boundaries and Intermediate scenarios.** Assuming very energy efficient data center with a 1.04 PUE, below current best High-Performance Computers from TOP 500, and electricity decarbonization in line with Paris agreement, hardware efficiency should be improved by a tremendous factor ranging from 175 to 565. This **demonstrates the need for new inference chip technologies that have ultra-low energy consumption.**

| | Energy usage | GHG Emission | Water consumption | PUE | Energy mix decarbonation factor | Hardware efficiency improvement factor |
|---|---|---|---|---|---|---|
| *Indexed 2024 portfolio* | 100 | 100 | 100 | | | |
| **High adoption without boundaries** | 755 | 576 | 650 | 1.15 | 0.75 | 4.4 |
| **Offset scenario** **(90% GHG reduction)** | 17 | 10 | 14 | **1.04** | **0.55** | **565** |
| **Intermediate scenario** | 2440 | 1862 | 2102 | 1.15 | 0.75 | 4.4 |
| **Offset scenario** **(90% GHG reduction)** | 17 | 10 | 14 | **1.04** | **0.55** | **175** |

## Discussion

The results of our portfolio analysis highlight critical considerations for the sustainable development of AI as a scaling technology.

### Transparency vs. Industrial Secrecy

The tension between transparency in AI's environmental impact and maintaining trade secrets is complex. Leading tech firms, driven by scalability and recovering their huge investments in Generative AI, often rely on proprietary AI training techniques for competitive advantage. Increased transparency could jeopardize this edge, discouraging full disclosure.

Revealing internal AI methodologies is crucial for environmental transparency but faces significant hurdles. Our approach investigates whether an AI system's environmental footprint can be estimated using a few publicly available parameters. This top-down method suggests reliable results with reduced complexity, encouraging wider corporate adoption.

Although our findings align with open-source models, validating this method for closed-source systems remains challenging. Future efforts should focus on creating a collaborative, open-access database involving tech stakeholders, certification bodies, NGOs, and governments to standardize practices while protecting proprietary information.

This is line with the EU AI Act that establishes that providers of general-purpose AI (GPAI) systems, that encompasses Generative AI models, should disclose information on the 'known or estimated energy consumption of the model' and documentation to improve resources consumption of AI systems during their lifecycle[36].

### Multi-criteria Environmental Footprint

Conventional life cycle assessments (LCA), using bottom-up methods, require specialized expertise and extensive data, limiting practical application. Our methodology strikes a balance between implementation effort and accuracy, supporting sustainable AI practices.

Moving beyond a sole focus on carbon emissions, it is crucial to assess potential trade-offs among various environmental impact components. The results reveal significant variation in the source of impact (e.g., operational versus embodied) depending on the factor considered. To capture the full spectrum of environmental consequences and minimize rebound effects on other impact areas, assessments should not be confined to a single factor such as greenhouse gas (GHG) emissions.

### Sustainability of Scaling Laws

Technological advancements and optimized AI usage have the potential to mitigate the escalating computational costs associated with larger models. Scaling laws suggest that performance improves through increasing model parameters and training data volume. However, for this paradigm to be economically viable, it presumes diminishing marginal costs, which depends on continued improvements in data center efficiency (PUE, WUE), decarbonization of electricity, and hardware optimization (CPU, GPU, TPU).

Despite notable efficiency improvements over the past decade, the long-term sustainability balance between scaling laws and technical progress remains uncertain. Our modeling, based on ceteris paribus ratios, provides partial insight into whether future innovations can maintain a sustainable AI trajectory in line with a 90% GHG reduction compared with 2024.

Based on our scenario *high adoption without boundaries*, our model suggests that efficiency improvements of 565x would be required to meet a net-zero target for AI by 2030. Current projections suggest that even the most optimistic efficiency gains described in our *Technological breakthrough* scenario fall short, heightening concerns over the long-term sustainability of scaling laws.

Only hardware technology major breakthroughs and large scale deployment, such as Neuromorphic computing or Cerebras' Wafer Scale Engines, could sustain scaling trajectories long-term. However, given their current technological readiness level (TRL), such innovations remain speculative for the near future[37].

## Can AI Models Be Scored Fairly?

Drawing inspiration from eco-scores in construction and textiles, an environmental scoring label for AI should balance precision and readability to reach a very large audience. A similar initiative has already been ideated by the community.[38,39] The scoring system could either focus on specific metrics such as energy consumption, carbon emissions, and water usage or aggregate these into a single score for simplicity, similar to France's DPE system that scores buildings energy performance. However, aggregation may compromise clarity due to the multidimensional nature of environmental impacts. Beyond usage, which is intuitive for the general public, incorporating factors like model training, semiconductor production, and end-of-life phases might improve accuracy but could also introduce complexity and uncertainty. Finally, defining AI environmental class thresholds could be based on either statistical distribution of current AI models (1) like in Energy Star project of Luccioni and al.[39] or the French Agency for Ecological Transition (ADEME) car environmental scoring[40] or, otherwise, a planetary boundaries approach aligning with global carbon budgets and finite resources (2)[41].

A critical aspect of energy evaluation in AI systems is the precise definition of both the scope and methodology for an energy score. A central question arises: should differentiation occur at the task level or model level?

On one hand, since large language models (LLMs) are employed across a wide range of tasks such as conversational agents, RAG, and agents, an energy score based on unitary actions (e.g., energy per token generated) could offer a normalized comparison of models. On the other hand, a broader comparison encompassing the entire spectrum of AI models—including computer vision, image generation, and time series forecasting—necessitates energy scoring based on tasks rather than models. However, defining a theoretical task for fair comparison is challenging. The work of Luccioni et al.[9,39] and Tschand and al.[42] provide foundational insights in this area defining key categories. However, with the rise of increasingly complex generative AI workflows, such as RAG and multi-agent systems, it is essential to continuously refine and discuss the definition of a "task" in energy rating systems. Enhanced granularity, especially for energy-intensive tasks, can improve the precision and fairness of energy impact assessments across diverse AI applications.

Table 6 presents a simplified example of threshold setting suggested by our work, employing the first method and based on a logarithmic difference between each threshold.

**Table 6 - Simplified AI environmental scoring.** Definition of an AI eco-score for the energy consumed per inference based on statistical distribution of current AI inferences.

| Naive AI eco-Score | A | B | C | D | E | F | G |
|---|---|---|---|---|---|---|---|
| **Energy per task (kWh)** | $< 10^{-8}$ | $< 10^{-7}$ | $< 10^{-6}$ | $< 10^{-5}$ | $< 10^{-4}$ | $< 10^{-3}$ | $< 10^{-2}$ |

## Towards an AI Return on Environment (RoE) Metric

Beyond direct environmental impacts (attributional approach), it is crucial to incorporate indirect effects (scope 4 / consequentialist approach) into AI sustainability assessment. Currently, the lack of a standardized framework and the difficulty of manually constructing counterfactual scenario for consequentialist methodologies hinder the ability to account for AI's potential positive indirect impacts on the environment. Enhancing existing methodologies is essential to guide the development of AI technologies toward a more responsible use, positioning AI as a positive force in climate change mitigation.

In conclusion, the environmental impact of generative AI depends on the responsible collaboration of various stakeholders. Three main factors influence its environmental footprint: widespread adoption across industries, increasing complexity of AI models and frameworks (particularly Agents), and the trend toward larger models. To minimize environmental risks, all actors of the AI value chain, including hardware manufacturers, must actively contribute to responsible deployment and usage. Achieving success requires greater transparency through information sharing among stakeholders, including environmental impact data and optimization methods. Without coordinated effort from model providers to end users, environmental impacts will significantly increase.

Future research should address several key areas:

- Integrating pre-training impacts, especially of Large Language Models, into value chain analysis to better attribute environmental costs across different enterprise applications.
- Continuously improving modeling methodology, especially in terms of parameter estimation and embodied emissions projections. This includes refining our understanding of hardware lifecycle impacts and improving energy consumption predictions for emerging AI architectures. To improve this framework, a more granular segmentation of typical industry use cases and model sizes could offer deeper insights.
- Expanding research beyond data center energy usage to examine environmental footprint of devices gathering data to be used by AIs especially IoT devices that are continuously gathering larger amount of data or end-user devices used to access AI services such as smartphones, watches or glasses.
- Initiating a dialogue with the community (companies, researchers) to further refine benchmarks and define "conventional" tasks, building on currently developed use cases within companies.

We encourage the broader research community and industry stakeholders to further investigate this topic through developing standardized measurement frameworks, creating open datasets documenting environmental impacts, and establishing collaborative initiatives.

# Methods

## Overall methodology

This paper aims at estimating the overall environmental impacts of a typical world top 2000 revenue company due to their AI systems. We define an approach (Figure 3) modeling the environmental impact at single use case level that will be aggregated to represent a company portfolio and project towards possible AI and global evolution trends in 2030. The developed approach is easy to maintain and relevant for industry experts without an in-depth expertise in AI or GenAI. It aims to provide insights, identify hotspots, and observe trends to enable effective eco-design actions and levers to limit impact.

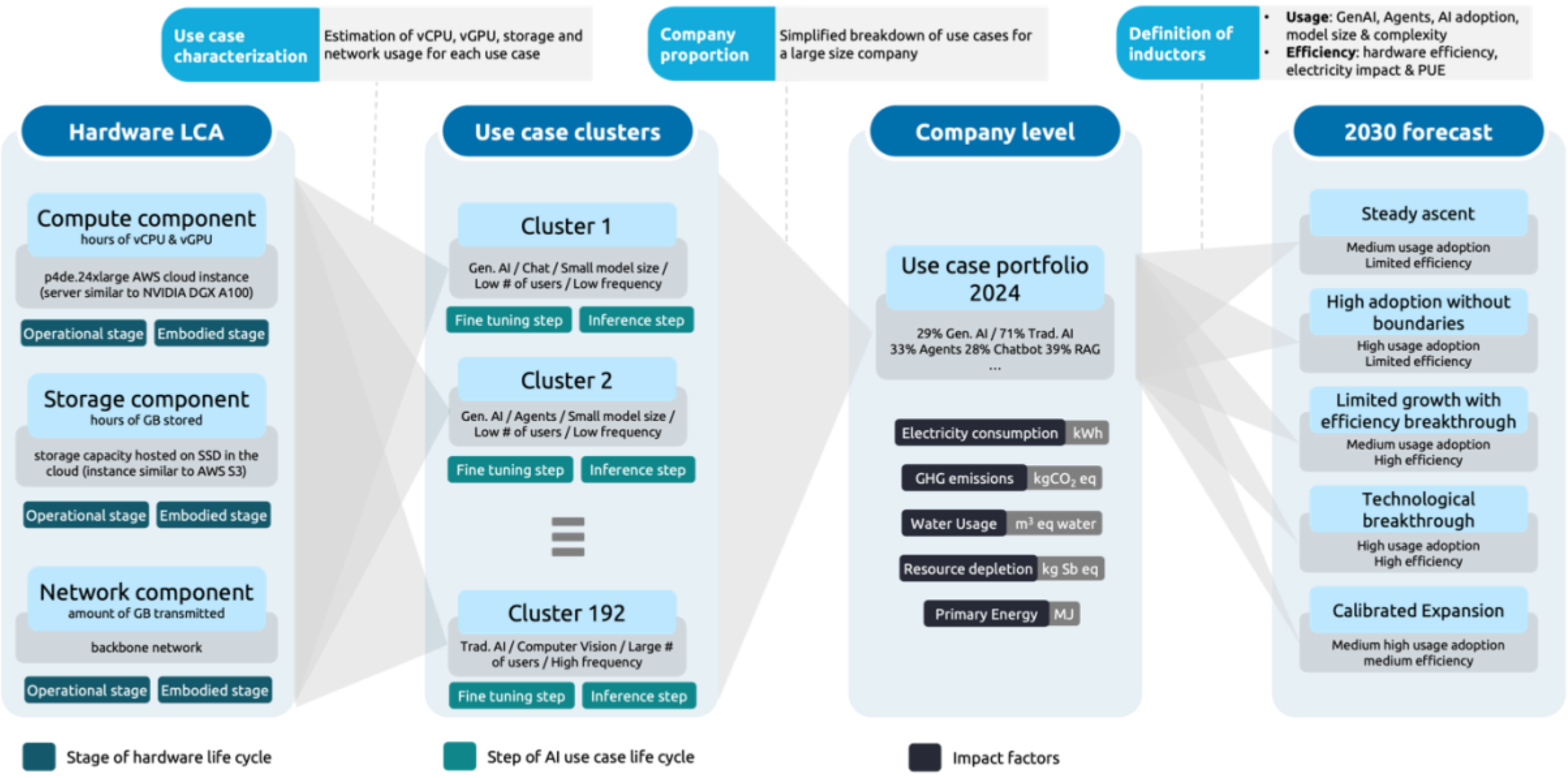

**Figure 3 – Methodology for assessing the environmental impacts of AI.** Our methodology leverages 4 sub models:

1. Life Cycle Assessment (LCA) model: an evaluation of energy and hardware usage life cycle footprint from production to end of life through several environmental impacts (aligned with PEF European recommendations) such as GHG emissions, water consumption or resource depletion.
2. AI Use case clustering: we define an AI clusters model based on criterions observed in various industry AI projects and estimate energy consumption and hardware usage for each cluster.
3. AI portfolio model: we create a fictive company AI portfolio to evaluate its AI related environmental footprint.
4. 2030 AI landscape model: Our methodology involves modeling various scenarios based on usage growth and systemic efficiency. We formulate 4 boundaries scenarios describing the extreme global state in 2030 regarding AI/GenAI adoption as well as efforts to reduce data centers footprint. Then, we perform sensibility analysis of various usage and efficiency parameters on an Intermediate scenario and the impact they would represent. Finally, we explore the efficiency needed to match a 90% GHG reduction compared with 2024.

**These models are merged into an excel simulator for analysis.**

## Model 1: Life-cycle-assessment methodology of hardware impacts

This study has relied on the Life Cycle Assessment (LCA) methodology to assess the impact of AI models. The key features of LCA are its multi-criteria, multi-step, and multi-component analysis.

The multi-criteria analysis considers five environmental indicators as follows: the global warming potential though the greenhouse gases emissions in kilograms of CO2 equivalent analyzed using the 100-year time horizon based on IPCC scientific report of 2013[43], the non-renewable abiotic resources depletion in kilograms of Antimony equivalent (kg Sbeq) analyzed using VanOers et al as in CML 2002 method (v4.8), the primary energy consumption in megajoules of primary energy (MJ) analyzed using VanOers et al as in CML 2002 method (v4.8)[44], the deprivation weighted water consumption in cubic meter of water analyzed using the available water remaining (AWARE) model[45] & the final energy consumption in kWh of electricity.

The multi-step analysis considers both the main steps of the AI use cases' life cycle & the hardware and associated infrastructures' life cycle. The AI project life cycle steps considered are fine-tuning and inference, while the hardware and infrastructures life cycle stages include the embodied phase, which involves the extraction of raw materials and manufacturing of equipment, and the operational phase, which involves the consumption of electricity and water during the use phase of equipment (servers) and infrastructures (datacenters, telecommunication networks).

The multi-component analysis assesses the complex system of components involved in AI use cases, defined in three capacity types: compute, storage, and network. Compute capacity is provided by CPU (central processing unit) and GPU (graphic processing unit) hosted in the cloud, expressed in hours of virtual CPU or virtual GPU. Storage capacity is expressed in hours of gigabytes stored, and network capacity is expressed in gigabytes transmitted from end-user devices to the hosted model in the cloud.

## Perimeter

The perimeter included the fine-tuning and inference steps of AI projects. On components, compute and storage capacities hosted on servers within public cloud datacenters, and communication between users and AI models on backbone networks are considered. The life cycle stage of equipment includes the embodied and operational stages.

## Impact evaluation

The environmental impacts of each use case are assessed using the formula below and its parameters are listed in Table 7 and sources for methodology in Table 8.

$$\boldsymbol{Impact_{total}(X_p, Sol_i) = \sum_{k=1}^{2}\sum_{n=1}^{4}\sum_{j=1}^{2} Impact(X_p, Sol_i, Step_k, Component_n, Stage_j)}$$

**Table 7 - Variables of the environmental impact formula.** List of the five parameters considered to assess the environmental impacts of each use case with their relative definitions and variables.

| Parameter | Definition | Variable |
|---|---|---|
| $\boldsymbol{X_p}$ | Environmental criteria | $X_1$: GHG emission (kg CO2 eq)<br>$X_2$: Non-renewable abiotic resources depletion (kg SB eq)<br>$X_3$: Primary energy consumption (MJ)<br>$X_4$: Deprivation of water consumption (m3 eq)<br>$X_5$: Final energy consumption (kWh) |
| $\boldsymbol{Sol_i}$ | Use case | Artificial intelligence solution 1 to 192 defined by specific parameters |
| $\boldsymbol{Step_k}$ | AI project life cycle step | $Step_1$: Fine tuning<br>$Step_2$: Inference |
| $\boldsymbol{Component_n}$ | IT capacity involved | $Component_1$: Compute (vCPU)<br>$Component_2$: Compute (vGPU)<br>$Component_3$: Storage (h.GB)<br>$Component_4$: Network (GB) |
| $\boldsymbol{Stage_j}$ | Life cycle stage of components | $Stage_1$: Embodied<br>$Stage_2$: Operational |

**Table 8 – Power and Embodied impacts for each IT capacity, Electricity emission factors for the USA, China & the EU & Water emission factors of the EU.** Power IT and embodied impacts of compute, storage, and transmission IT capacities. Operational impacts are evaluated by applying to the power IT a PUE of 1.15 and a WUE of 0.18 L/kWh, using an electrical mix from the USA, China, and the EU, and a European water mix.

Each assessment relies on the emission factors databases of the hardware, meaning the embodied and operational impacts of the compute, storage and network capacity. The construction of this database has been realized using the representative components. The p4de.24xlarge38[46] which is similar to NVIDIA DGX A10039 hosted on AWS cloud was considered on compute to model both the usage of one virtual CPU and one virtual GPU for one hour. The storage instance using a server associated with a storage bay hosted on AWS was considered on storage to model the storage of one GB of data for one hour. An extrapolation of the European backbone network was performed to model the transmission of 1 GB on the backbone network. A PUE of 1.15 and a WUE of 0.18 L/kWh were also considered on compute and storage.

The embodied impacts have been estimated using a bill of material approach based on the configurations of the reference's equipment mapped to LCA data in the NegaOctet database and associated with the capacity of each equipment. A standard lifetime of 4 years has been considered as hypothesis. The operational impacts have been evaluated using the location-based method to calculate emissions based on the average emissions intensity of the local electricity grid where the electricity is consumed.

The NegaOctet database has been used to model the environmental impact of each equipment and to define the following impact tables used for the assessment of each use case.

| | Power IT | PUE of datacenter | WUE | GWP | WU | TPE | ADPe |
|---|---|---|---|---|---|---|---|
| **1h vGPU** | 5,01E+01 W | 1,15 | 0,18 L/kWh IT | 1,93E-03 kgCO2eq/h | 6,59E-04 m3eq/h | 2,85E-02 MJ/h | 9,84E-09 kg SBeq/h |
| **1h vCPU** | 3,15E+00 W | 1,15 | 0,18 L/kWh IT | 1,67E-04 kgCO2eq/h | 5,34E-05 m3eq/h | 2,49E-03 MJ/h | 3,85E-08 kg SBeq/h |
| **1h storage 1GB** | 1,25E-03 W/GB | 1,15 | 0,18 L/kWh IT | 1,11E-06 kgCO2eq/(h.GB) | 4,81E-07 m3eq/(h.GB) | 4,95E-06 MJ/(h.GB) | 1,31E-11 kg SBeq/(h.GB) |
| **Transmission 1GB backbone** | 3,42E-02 Wh/GB | NA | NA | 3,65E-04 kgCO2eq/GB | 1,17E-04 m3eq/GB | 5,65E-03 MJ/GB | 5,85E-08 kg SBeq/GB |

| Regional electricity emission factor | GWP (kg CO2 eq/kWh) | Water use (m3 eq/kWh) | Primary energy use (MJ/kWh) | Resource use (kg SB eq/kWh) |
|---|---|---|---|---|
| **USA** | 5,47E-01 | 1,86E-02 | 1,16E+01 | 2,21E-08 |
| **China** | 8,71E-01 | 3,82E-02 | 1,56E+01 | 1,12E-08 |
| **EU-27** | 4,10E-01 | 1,36E-02 | 1,25E+01 | 2,97E-08 |

| Regional Water emission factor | GWP (kg CO2 eq/L) | Water use (m3 eq/L) | Primary energy use (MJ/L) | Resource use (kg SB eq/L) |
|---|---|---|---|---|
| **EU-27** | 5.84E-04 | 4.31E-02 | 2.42E-03 | 6.28E-10 |

## Model 2: Impact of AI use cases

We developed a comprehensive methodology for estimating the energy consumption and its resulting environmental impact of AI solutions considering the power consumption of CPUs and GPUs, data storage requirements, and network data transmission. This detailed approach leverages technical specifications, usage rates, and various assumptions to provide a thorough analysis. For an in-depth understanding of the underlying assumptions, data sizes, and specific modeling techniques, please refer to the supplementary material sections on AI model Hypothesis, Energy breakdown (Compute, storage, network), Inference & Fine tuning.

To facilitate the modeling of a typical company portfolio, use cases were categorized into distinct clusters based on five dimensions. The first dimension differentiates between Generative AI and Traditional AI. For Generative AI, use cases include Chat, Retrieval-Augmented Generation (RAG), or Agent-based applications, while for Traditional AI, use cases encompass Tabular data analysis, Computer Vision, or Natural Language Processing (NLP). Model size, applicable only to Generative AI, is classified into three categories based on actual model distributions. The number of daily users is represented on a logarithmic scale, ranging from 10 users for Proof-of-concept projects, 100 for Minimal viable products, 1000 for industrialized projects and 10000 for widely scaled use cases (internal chatbots, Copilot, …). This resulted in 192 unique clusters with more Generative AI clusters due to model size variations. This framework simplifies adoption patterns and future projections*.*

## Model 3: Company portfolio model

In this context, we modeled a typical company's AI portfolio to align with the proposed AI clusters.. We defined our model with 4 main components representing the main axes of the AI clustering model:

- Ratio of Traditional AI vs Gen AI use cases
- Ratio of use case type along the proposed categories: tabular, computer vision, natural language processing, chatbot, RAG and AI agents
- On AI and GenAI use cases, the ratio of users' number and usage frequency along the 4 proposed categories (low, medium, high, very high)
- Ratio of GenAI model size along the 3 proposed categories (low, medium, high)

Assuming uniform distribution, the ratios are multiplied to compute the overall ratio of use case per cluster.

The proposed portfolio will be based on a company with 20+ billion annual revenue. Those companies are usually more mature in AI adoption. According to Capgemini Research[47], 49% of which have already implemented GenAI solutions and 89% have increased their investments over the past year.

Details about each ration modeling is presented in supplementary information in section Company Portfolio Model.

## Model 4: Projected 2030 scenario

To project our company portfolio, we chose to use 2030 at reference. This choice is based on 2 motivations. First, it is in line with various climate and energetic scenarios and targets such as the ones from IPCC reports on energy systems[48], or the European Union[49] or AIE World Energy outlook[35]. Secondly, most of market analysis of Artificial intelligence and Generative AI use 2030 as reference as well. This will be helpful to project the usage and distribution of our company.

We have chosen to model various scenarios based on 2 levers: usage projection (number of use cases, distribution) and Systemic efficiency projection (size & complexity of models, compute efficiency, electricity impact…). The methodology follows a three-step process:

- Establishment of **boundary scenarios** to understand the maximum impact range of usage development, combining minimum and maximum bounds under a "no intervention" assumption, alongside an idealized scenario of global efficiency improvements.
- Development of **intermediate scenarios**, we explore the impact of various parameters on an Intermediate scenario. We try and reflect anticipated usage patterns, efficiency gains in computational performance and resource utilization to explore the potential scaling factors they could induce.
- Analysis of these scenarios to evaluate the **potential impact** of different technologies and identify strategies to mitigate the explosive growth in generative AI usage.

### Boundaries scenarios

**Usage adoption:** This involves defining a lower and upper bound for the penetration of generative AI, AI models, and autonomous agents. Projections include model size, complexity, and hardware energy efficiency, coupled with electricity grid emission forecasts. The two scenarios modeled are "*Steady Ascent*" and "*High adoption without boundaries*" reflecting the potential impact without further technological improvements or usage constraints.

**Systemic Efficiency:** These scenarios represent theoretical extremes of computational efficiency and resource frugality, where technological solutions drive major efficiency gains. They explore the outcomes if sustainable AI usage is prioritized, IPCC electricity targets are met, and compute technologies are significantly optimized. On the one hand, "*Limited growth with efficiency breakthrough"* follows a steady ascent usage growth coupled with frugal development and usage of LLMs and hardware efficiency breakthrough while "*Technological breakthrough*" mainly relies on hardware efficiency.

### Intermediate Scenarios:

Building on these boundary cases, we have performed various sensibility analysis of usage and efficiency parameters on an *Intermediate scenario*. We aim to assess the potential scaling impact of growing model sizes and Agentic system adoption. Finally, we have explored the hardware efficiency gains required to achieve a 90% GHG reduction compared with 2024.

Our *Intermediate* scenario, following conservative trends in energy efficiency improvements and hardware power density as well as median adoption models.

Based on this intermediate scenario, we have explored the sensitivity of various parameters to assess the potential impact their variations might represent

- Model size: evolution of sizes of LLMs.
- Agentic systems deployment based on their level of penetration and complexity
- Hardware efficiency necessary to achieve substantial reduction of impacts.

For details about scenario hypothesis see the sections 2030 Systemic projections and 2030 Systemic projections of supplementary information.

A detailed summary of scenarios is shared in the supplementary information Summary of scenarios

## Limits of the methodology

The modeling approach considered presents certain limitations.

- Predefined clusters may oversimplify nuanced or hybrid applications, missing industry-specific constraints. The framework offers high-level projection-making but requires qualitative insights or custom analyses for specific cases. Improvements include:
    - Task Segmentation: Refining NLP and CV task analysis with detailed data.
    - Model Size Segmentation: Analyzing how model complexity affects energy and resources. For instance, based on model performance, the number of output tokens might defer, a more granular analysis might enhance this analysis.
- The statistical representation employed for our typical company AI portfolio, while effective for projecting impacts in 2030, may oversimplify the complexity of real-world usage scenarios. To enhance the methodology, we recommend that companies and researchers refine the statistical distribution by adopting a statistical usage by task. Additionally, we suggest leveraging real-world AI portfolios and mapping them to the 192 clusters provided in our framework for a more comprehensive and realistic assessment.
- A balanced and comprehensive selection of data sources is critical to mitigate the risk of bias in the analysis. This could arise from inconsistencies in the analysis, such as hardware configurations of providers used for latency calculations by Artificial Analysis[31]. On the other hand, relying on too few data sources, for example for estimating the sizes of requests could lead to incomplete or skewed representations, limiting the model's ability to provide accurate and generalizable insights.
- Transparency: Some data are currently unavailable and are therefore estimated empirically[9,31]. To improve accuracy in energy consumption estimates, it is crucial for model providers to release more detailed information about their models' architectures and theoretical energy requirements. This transparency would enable the use of more precise estimation methods.
- Pre-training impact: Our analysis emphasizes the importance of considering all phases of a model's life cycle, including pre-training, continuous training, and end-of-life management. However, due to limited transparency on the distribution of general public usage versus corporate-specific applications, it remains challenging to fairly allocate the environmental impact of pre-training across all users. Further research on statistical modeling of global usage patterns could refine this aspect of the analysis.
- Life Cycle Assessment (LCA) methodology: LCA methodology can either be attributional (focusing on current impacts) or consequential (considering system-wide effects)[50]. This study concentrates on the attributional part of AI impacts and future work should address the consequential LCA. Only a coupled attributional and consequential approach would lead to assess the overall usefulness of an AI system[52] especially when it comes to AI for green.
- 2030 Projections and Uncertainty: The model incorporates multiple projection factors that may evolve significantly and are subject to considerable uncertainty. Therefore, it is essential to discuss both the methodology and the rationale behind the selection of these parameters. A more granular approach, including projections of embodied rebound effects for new data center infrastructure and differentiated impact factors for electricity grids, could further enhance the robustness of the analysis.
- Emission Factor Database: The accuracy of emissions estimates is directly influenced by the choice of emission factor databases. Differences in geographic contexts, data freshness, and methodological assumptions across databases can affect the results. Future iterations could benefit from a standardized, fine grained and regularly updated database.

- Impact Indicators Definition: The selected impact indicators, while informative, may not fully capture all dimensions of environmental impact. Expanding the range of indicators considered, such as water pollution, e-waste, or social impacts, could provide a more comprehensive assessment.

The results presented in this paper are based on parameters that, despite efforts to ensure rigor, remain subject to uncertainty due to limited data transparency and the inherent unpredictability of future projections. While the methodology aims to provide a robust assessment, these limitations may influence the interpretation of results and introduce potential biases. This approach is open to further refinements, and we encourage the broader research community to contribute to its continuous improvement, ensuring a more reliable assessment of the environmental impact of AI technologies.

## Acknowledgements

We would like to express our deepest gratitude to the direction of Capgemini Invent for their sponsorship and insightful reviews of the paper, with particular acknowledgment to Etienne Grass, Managing Director of Capgemini Invent France, and Philippe Cordier and Nicolas Brunel, Director and scientific Director of the Research & Innovation Lab of Capgemini Invent France.

We are also profoundly grateful to the members of the Capgemini Invent Lab for their expert guidance throughout the submission process and their constructive feedback on the article. Their assistance has been pivotal to the successful completion of this work.

Furthermore, we extend our heartfelt thanks to our colleagues at Capgemini Invent, whose expertise and stimulating discussions have greatly enriched this research. We would like to give special recognition to Anh Khoa Ngo Ho for his dedicated support and insightful contributions to the related studies.

This work would not have been possible without the collective effort and collaboration of all involved, and we are deeply appreciative of their dedication and commitment.

## Author contributions

All the listed contributions are based on alphabetical order. M.C., P.C. and S.G. designed and planned the project. C.D., C.V., L.L. and M.C. designed the Life cycle assessment perimeter and global methodology. C.V. and L.L. developed the methodology to assess servers' multi-factor impact. C.D., M.C. and P.C. developed the methodology to calculate AI use cases consumption. C.D., L.L., C.V, M.C., S.G. and P.C. designed the 2030 scenarios and projections. C.D. and M.C. performed the analyses and wrote the associated results with feedback from C.V., S.G. and P.C., C.D., C.V., L.L., M.C. and S.G. wrote the manuscript and incorporated feedback from other authors.

## Competing interests

The authors are either employees, shareholders, or stock option holders of Capgemini.

# Supplementary information

## LCA indicators

**Table 9 - LCA indicators.** List of the five LCA indicators considered in the study with their characteristics.

| Type of indicator | EF Impact category | Impact category indicator | Impact category description | Unit | Characterization model | Robustness (EF level)[53] |
|---|---|---|---|---|---|---|
| **Impact** | **Climate change, total** | Global warming potential (GWP100) | Most known indicator, refers to the modification of climate impacting the global ecosystem. It is the potential global warming resulting from GHG emissions into the atmosphere.<br>There are three subcategories based on the source of emission: fossil fuel, bio-based resources, land use / land use change. All GHGs covered by the Kyoto Protocol / UNFCCC. | kg $CO_2$ -eq | Bern model - Global warming potentials (GWP) over a 100-year time horizon (based on IPCC 2013) | I |
| | **Water use** | User deprivation potential (deprivation weighted water consumption) | assesses the total volume of water consumed or withdrawn for various activities within a system or process. It accounts for both direct water consumption and indirect water use, such as water embedded in products or services. This indicator helps quantify the impact of water resource utilization associated with a particular activity or process. | $m^3$ eq. water<br>($m^3$ water eq of deprived water) | Available Water REmaining (AWARE) model (Boulay et al., 2018; UNEP 2016)[45] | III |
| | **Abiotic resources use – minerals and metals** | Abiotic resource depletion (ADP ultimate reserves) | Indicator of the depletion of natural **non-fossil resources**: copper, potash, rare earths, sand, etc. | kg Sb-eq | van Oers et al., 2002 as in CML 2002 method, v.4.8[44] | III |
| | **Primary energy use (Resource use – fossils)** | Primary energy use – fossil fuels (ADP-fossil) | Indicator of the depletion of natural **fossil fuel resources**: coal, gas, oil, uranium, etc. | MJ | van Oers et al., 2002 as in CML 2002 method, v.4.8[44] | III |
| **Flow data** | **Final energy use i.e., annual electricity consumption** | | Measures the total energy consumption at the end-use stage, reflecting the energy demand of a system or process after considering losses in distribution, conversion, and utilization. It quantifies the energy required for various activities, including heating, cooling, lighting, and appliance operation. | kWh | | N/A |

# LCA Methodology

## Compute model

For the **embodied impacts assessment**, a bill of material methodology has been used considering the following data, hypothesis and sources as the environmental data have not been published.

**Table 10 - Specifications of the compute model.** The compute model included electronic cards for CPU & GPU, CPU & GPU chips, power supply, casing, fan, mother board, disk and RAM components. Operational & embodied impacts are allocated to vCPU & vGPU to define both the vCPU and vGPU functional units.

| Technical specifications | Material for LCA | Emission Factor Considered | Allocation to IT capacity |
|---|---|---|---|
| NVDIA A 100 Tensor core 80 Gb SXM | 2.96x10-2 m2 of electronic card*<br>4.83x10-2 m2 of Wafer, 75 die, 7nm EUV lithography** | NegaOctet | 100% to vGPU |
| CPU Xeon Platinum[46]CPU Xeon Platinum[46] | 2.01x10-1 m2 of electronic card*<br>4.31x10-3 m2 of Wafer, 58 die, 14nm EUV lithography** | NegaOctet | 100% to vCPU |
| Power Supply | | NegaOctet | For each environmental indicator:<br>$\% \, allo \, vCPU = \frac{Impact \, vCPU}{Impact \, (vCPU + vGPU)}$<br>$\% \, allo \, vGPU = \frac{Impact \, vGPU}{Impact \, (vCPU + vGPU)}$ |
| Casing | | | |
| Fan | | | |
| Mother board | 4.33x10-1 m2 mother board | | |
| Disk | 8 SSD, 1024 GB | | |
| RAM | 4 Random Access Memory DDRS 256 GB | | |

** Assessed in section Electronic card impacts*

*** Assessed in section Chips impacts*

For the operational impact assessment, the electricity consumption of the hardware system has been assessed to 3,110W without additional impact due to the hosting within datacenter and to 4,665W including the PUE. The specification of the hardware and the following calculation rule were used.

$$\begin{aligned} \boldsymbol{P_{compute\ model}(W)} &= \boldsymbol{N_{CPU} * \Big(Min(P_{CPU}) + Load\ rate\ CPU * \big(Max(P_{CPU}) - Min(P_{CPU})\big)\Big)} \\ &\quad \boldsymbol{* (1 + \%orchestrator\ ) + N_{disk} * P_{disk} * Load\ rate\ disk * Replication} \\ &\quad \boldsymbol{* (1 + \%add_{disk}) + N_{GPU} * Load\ rate\ GPU * Max(P_{GPU}) * (1 + \%add_{GPU})} \\ &\quad \boldsymbol{+ \ N_{RAM} * Load\ rate\ RAM * P_{RAM} * (1 + \%add_{RAM})} \end{aligned}$$

The same allocation rule used for embodied emission has been used based on vCPU and vGPU embodied emissions.

**Table 11 - Compute model parameters.** List of the parameters considered to assess the power of the compute model.

| Parameter | Value |
|---|---|
| $N_{GPU}$ | 8 |
| $Load\ rate\ GPU$ | 80% |
| $Max(P_{GPU})$ | 400W |
| $\%add_{GPU}$ | 5% |
| $N_{CPU}$ | 2 |
| $Load\ rate\ CPU$ | 50% |
| $Max(P_{CPU})$ | 240W |
| $Min(P_{CPU})$ | 35.52 W |
| $\%orchestrator$ | 5% |
| $N_{disk}$ | 24 |
| $P_{disk}$ | 18W |
| $Load\ rate\ disk$ | 80% |
| $Replication$ | 3 |
| $\%add_{disk}$ | 5% |
| $N_{RAM}$ | 4 |
| $Load\ rate\ RAM$ | 50% |
| $P_{RAM}$ | 8.5W |
| $\%add_{RAM}$ | 5% |

## Storage model

For the **embodied impacts assessment**, as the environmental data have not been published, a bill of material methodology has been used considering the following data, hypothesis and sources.

**Table 12: Specifications of the storage model.** The storage model includes electronic cards CPU & its chips, power supply, casing, fan, mother board, disk and RAM components. Operational & embodied impacts are allocated to storage to define the storage functional unit.

| Technical specifications of the storage system | Material for LCA | Emission Factor Considered | Allocation to IT capacity |
|---|---|---|---|
| **CPU Xeon Platinum** | 2.01x10-1 m2 of Electronic card<br>4.31x 10-3m2 of Wafer, 58 die, 14nm EUV lithography | NegaOctet | 100% to hour.GB |
| **Power Supply** | | | |
| **Casing** | | | |
| **Fan** | | | |
| **Mother board** | 4.33x10-1 m2 mother board | | |
| **Disk** | 8 SSD, 15To (3 replications) | | |

For the operational impact, the electricity consumption of the system has been assessed to 1,378W without additional impact due to the hosting within datacenter and to 1,583W including the PUE. The specification of the hardware and the following calculation rule were used.

$$P_{storage\ model}(W) = N_{CPU} * \left(Min(P_{CPU}) + Load\ rate\ CPU * \left(Max(P_{CPU}) - Min(P_{CPU})\right)\right) * (1 + \%orchestrator) + N_{disk} * P_{disk} * Load\ rate\ disk * Replication * (1 + \%add_{disk})$$

**Table 13 - Storage model parameters.** List of the parameters considered to assess the power of the storage model.

| Parameter | Value |
|---|---|
| $N_{CPU}$ | 2 |
| $Load\ rate\ CPU$ | 50% |
| $Max(P_{CPU})$ | 240W |
| $Min(P_{CPU})$ | 35.52 W |
| $\%orchestrator$ | 5% |
| $N_{disk}$ | 24 |
| $P_{disk}$ | 18W |
| $Load\ rate\ disk$ | 80% |
| $Replication$ | 3 |
| $\%add_{disk}$ | 5% |

## Electronic card impacts

The impact of 1 m² of electronic card is known using the manufacturing code NEGA-0052 with “Motherboard; mix of equipment, without processor or RAM” as component name. Therefore, the surfaces of electronic cards hosting GPU & CPU chips are evaluated.

At first, 8 GPU chips are hosted on the electronic card PCI-Express 4.0 x16[54] as this card is used for a A100 PCIe 80 GB server that could be utilized to run the p4de.24xlarge instance. This card measures 267 mm in length & 111 mm in width which leads to a surface of 2.96E-02 m².

Moreover, 2 CPU Intel Xeon Platinum 8275CL[55] used for p4de.24xlarge has a TDP of 240W[56] and 24 cores. Each CPU is considered hosted on the electronic card X12SPA-TF[57] supporting TDP up to 270W & up to 40 cores. This card measures 33.02 cm length and 30.48 cm width which leads to a total surface of 2.01E-01 m² considering the surfaces of 2 electronic cards.

## Chips impacts

CPU and GPU chips are made from silicium wafers. Wafers undergoes 3 types of significant losses before obtaining CPU & GPU chips (see figure below).

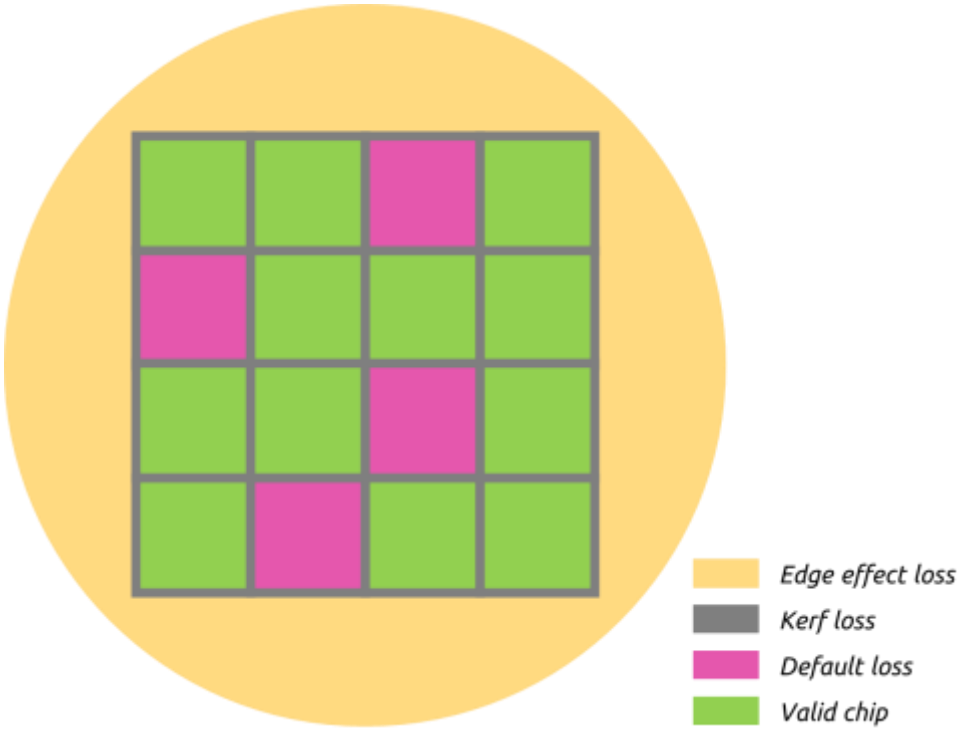

**Figure 4 - Losses undergoes by wafers to create CPU or GPU chips.** Chips undergoes edge effect, kerf & default lossed during their manufacturing process.

The area of silicium needed to create a chip considering these losses is evaluated to assess its environmental impacts using the manufacturing code of wafers, where $A_{chip}$ is the area of a chip, yield is the yield related to both the edge effect & kerf loss and $Y_{default\ loss}$ is the yield related to the default losses.

$$\boldsymbol{A_{silicium\ needed} = \frac{A_{chip}}{yield * Y_{default\ loss}}}$$

### Yield evaluation

The edge effect loss occurs when cutting square chips from circular wafers while the kerf loss refers to the material lost during the cutting process of silicon wafers. A first yield is used to consider the edge effect & default losses, where $N_{chip}$ is the number of chips to be created in a wafer and $A_{wafer}$ is the area of a wafer.

$$\boldsymbol{yield = \frac{N_{chip} * A_{chip}}{A_{wafer}}}$$

The largest wafer used to create chips has a diameter of 300mm[58], therefore $A_{wafer}$ is known. $A_{chip}$ of both the CPU & GPU chips are known based on their specifications.

### Evaluation of $N_{chip}$

The evaluation of $\boldsymbol{N_{chip}}$ is known using the following formula[59], where $D_{wafer}$ is the diameter of the wafer and $A_{chip\ with\ kerf}$ is the area of chip considering its kerf.

$$\boldsymbol{N_{chip} = \frac{\pi * (\frac{D_{wafer}}{2})^2}{A_{chip\ with\ kerf}} - \frac{\pi * D_{wafer}}{\sqrt{2 * A_{chip\ with\ kerf}}}}$$

### Evaluation of $A_{chip\ with\ kerf}$

Chips are cut from the wafers with a width called the kerf (see figure below) and are considered a square for simplification purposes.

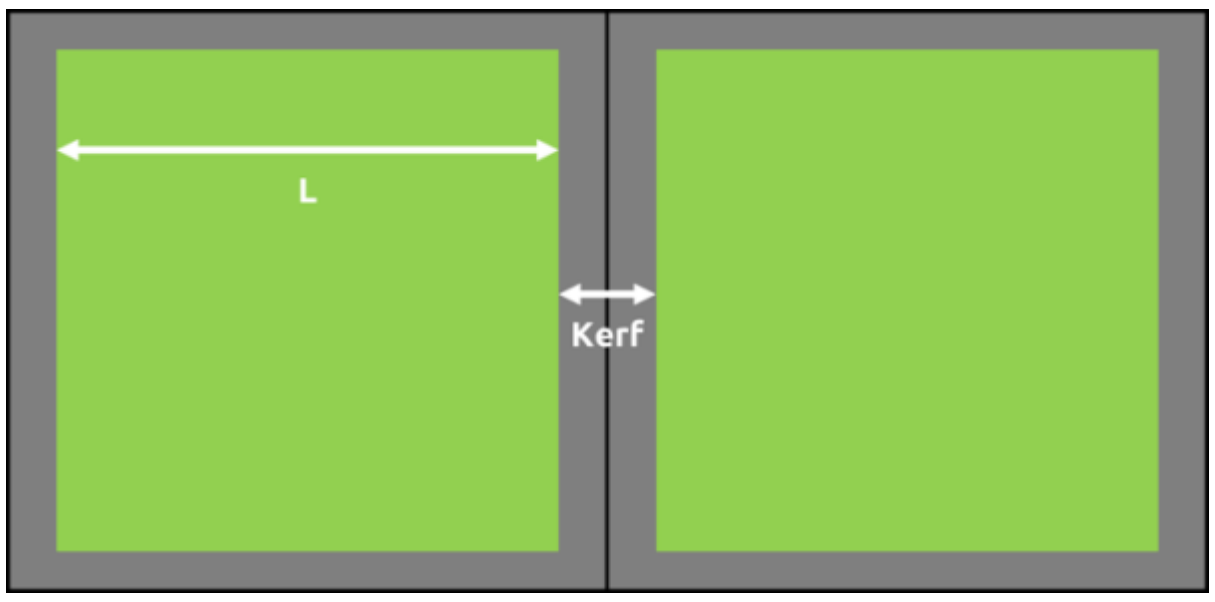

**Figure 5 - Identification of kerf in silicium cutting.** Cutting chips from wafers generate kerf losses.

The area $\boldsymbol{A_{chip\ with\ kerf}}$ of each chip is determined after considering the kerf, where L is the length of the final square chip & kerf is the width of the cut

$$\boldsymbol{A_{chip\ with\ kerf}} = (\boldsymbol{L} + \boldsymbol{kerf})^2$$

As the chip is considered a square, L is easily known where $A_{chip}$ is the area of the final chip.

$$\boldsymbol{L} = \sqrt{\boldsymbol{A_{chip}}}$$

$A_{chip\ with\ kerf}$ is evaluated as $A_{chip}$ was evaluated before and by considering a kerf of 0.2 mm[60].

## Evaluation of $Y_{default\ loss}$

Some defects on wafer lead to the failure of chips. These defects are considered with $Y_{default\ loss}$ which is given by the formula[61] below using the Moore model, where D is the defect density of the wafer.

$$Y_{default\ loss} = e^{-\sqrt{D * A_{chip}}}$$

The results below are obtained.

$$\boldsymbol{A_{silicium\ needed,\ \ CPU\ chip}} = \boldsymbol{4.31 * 10^{-3} m^2}$$

$$\boldsymbol{A_{silicium\ needed,\ \ GPU\ chip}} = \boldsymbol{4.83 * 10^{-2} m^2}$$

Finally, the impacts of both chips were evaluated considering their area of silicium needed and their manufacturing code.

## AI model Hypothesis

In this section, we detail all the analysis and hypothesis used in the calculation methodology, from AI training / inference to LCA impact of instances.

## Energy breakdown (Compute, storage, network)

In the following section, we'll detail the energy estimation methodology for a single AI use case. Embodied impacts are estimated similarly using hourly impact factors instead of power consumption.

Energy consumption estimation

The first step of our impact methodology is to estimate the energetic consumption of an AI solution both for its fine tuning and inference phases.

$$\boldsymbol{E_{solution_i} = E_{sol_i} Fine\ tuning + E_{sol_i} Inference}$$

Throughout these two phases, three types of theoretical energy consumption are involved based on technical specifications of components and usage rates.

- **Compute:** This refers to the electricity used by GPUs and CPUs to run the model effectively.
- **Storage**: This pertains to the electricity required to store data for training or to monitor inferences.
- **Network**: This involves the electricity consumed during the transmission of data between users' computers and the model server including sending the request and receiving the answer.

### Compute

To model compute usage of AI systems, we used a similar approach as Ecologits[25] calculator, dividing GPUs consumption and remaining server's one. We need to 2 items. First, the usage in total Hours of hardware. The total number of hours for training and Fine tuning on one side; and the inference latency on the other depending on the nature of the task (NLP / CV / Tabular / LLM) and the number of GPUs required to run the model (based on model size). Secondly, we need to estimate the power consumption of GPUs and servers used to run the model depending on the server instance considered.

$$\mathbf{E}\ \boldsymbol{Compute} = \mathbf{E}\ \boldsymbol{CPU} + \mathbf{E}\ \boldsymbol{GPU}$$

$$= \boldsymbol{vCPU\ hours} * P_{vCPU} * + \boldsymbol{vGPU\ hours} * P_{vGPU}$$

With:

$- vCPU\ hours : time\ (in\ hours)\ of\ vCPU\ usage\ to\ pre\ train, fine\ tune, infer\ the\ model$
$- P_{vCPU} : vCPU\ power\ (in\ W)\ of\ the\ cloud\ instance$

$- vGPU\ hours : time\ (in\ hours)\ of\ vGPU\ usage\ to\ pre\ train, fine\ tune, infer\ the\ model$
$- P_{vGPU} : vGPU\ power\ (in\ W)\ of\ the\ cloud\ instance$

For fine tuning, we detailed our usage estimation in the section Fine tuning below. It is mainly based on inference usage multiplied by the estimated number of operations (training batch size * epochs) for Traditional AI models considering most of companies' AI solutions are fine-tuned on existing pre trained models. Note that we do not consider any LLM fine tuning for the GenAI use cases considered.

For inference usage, it mainly relies on studies conducted by Luccioni et al[9] and Artificial Analysis[31] to get the energy of 1 inference, and we multiply by the number of inference yearly (250 business days * nb of users * usage frequency). The detailed hypothesis on model sizes / GPU needed / quantization / latency can be found in the section Inference of supplementary work.

The details to understand the LCA modeling of hardware components at vCPU and vGPU level is presented in the supplementary information in the section LCA Methodology.

### Storage

To estimate the required energy to store the data, we estimate 3 variables: the total amount of Data, the number of hours of storage of the data and the power consumption of a storage unit per Gb.

And the resulting energy would be:

$$\boldsymbol{E_{sol_i}Storage\ (Wh) = Size\ data\ (Gb) * P_{storage}(\frac{W}{Gb}) * Storage_{time}(hours)}$$

**For inference,** we consider the energy required to store the data used in all inferences during the whole year.

And the resulting size would be:

$$\boldsymbol{size\ data\ (Gb) = 250 *\ Nb\ of\ users * usage\ frequency * Size\ data\ per\ inference}$$

And storage time is 24*365 hours as we consider all inferences to be stored all year.

The details of size data per inference depends on the nature of the use cases. The details of average request size can be found in the section Inference of supplementary information.

**For the fine-tuning** phases, we made assumptions regarding the sizes of datasets used for specific use case fine-tuning. Since averaging dataset sizes for standard use cases is a challenging task, these assumptions remain open for further discussion and refinement.

To model $P_{storage}$, the AWS instance S3 was considered as a reference as it is frequently to store data. Details about the modeling of this instance is presented in section Storage of supplementary information.

### Network

Finally, we account the energy necessary to transit the data through network calls estimated simply as the size of data sent and received from an inference call or downloading the training dataset multiplied by the network impact in Wh/Gb.

And the resulting energy would be:

$$\boldsymbol{E_{sol_i}Network = Size\ data * E_{network}}$$

With:

$- Size\ data:\ in\ Gb, the\ amount\ of\ data\ used\ for\ training\ or\ for\ inference$

$- E_{network}: Energy\ \left(in \frac{kWh}{Gb}\right)\ to\ download\ or\ upload\ 1\ Gb\ of\ data$

For pre training and fine-tuning phases, we consider that we download the entire dataset size once. See the datasets considered in the supplementary information section Fine tuning.

For inference, we consider the energy required to send and receive the data used during all inferences yearly with upload considering input size and output size for the output. The details of size data per inference depends on the nature of the use cases. The details of average request size can be found in the supplementary information section Inference

The same hypothesis is used to estimate size of training and inference data.

## Inference

### Traditional AI

To estimate the usage of vCPUs and vGPUs for Traditional AI use cases, we based our analysis on the experimental results presented by Luccioni et al.[9]. In their study, the inference energy consumption of various standard AI tasks in Computer Vision (e.g., image classification and object detection) and Natural Language Processing (e.g., fill-mask, question answering, text classification, and token classification) was evaluated using the eight most popular models on Hugging Face for these domains. The experiments were conducted on an AWS instance equipped with eight NVIDIA A100-SXM4-80GB GPUs. We adopted the same instance setup to model the embodied impacts of servers.

We retrieved the experimental data from their GitHub repository and extracted the average energy consumption for Computer Vision (CV) and Natural Language Processing (NLP) tasks, distinguishing between CPU, GPU, and RAM usage. Using this data, along with the reported average power, we calculated the per-second resource usage of the instances. For clarity and simplification, we chose to group RAM and GPU energy together while treating CPU separately, as RAM usage was primarily influenced by model loading.

Note that, for our study, we consider the impact at the vCPU and isolated multi-instance vGPU level. A vCPU (Virtual Central Processing Unit) is a virtualized version of a physical CPU core allocated to a virtual machine (VM) in a cloud computing environment. It represents a portion of the physical CPU's resources. Similarly, MIG (Multi-Instance GPU), is a technology developed by NVIDIA that allows a single physical GPU to be partitioned into multiple independent instances, each acting like a smaller, isolated vGPU.

As explained in Life-cycle-assessment methodology, our instance is composed of 96 vCPU delivering each a power of 3.15 W and 8 * 7 isolated Multi instance vGPU with up to 10GB of memory delivering a power of 50,1W.

To determine the number of vCPUs and vGPUs used, we employed a cross-multiplication method based on the average power consumption reported in the experiments of [9] to align with their energy per inference measurements.

For standard tabular data use cases, we conducted a similar simulation considering a Random Forest Classifier model, for 50 different float features, running on eight CPUs and a RAM of 6W for a total power consumption of 48.5W (derived from Apple's M2 architecture). This configuration was used to approximate the number of virtual CPUs (vCPUs) in our instance setup. As a result, we get the following figures:

**Table 14 - Inference compute usage hypothesis - Traditional AI.** Energy consumption related to compute hardware considered during the inference for three traditional AI use cases associated to compute.

| Use Case type | Energy 1 inference (kWh) Compute |
|---|---|
| Tabular | 2,99E-08 |
| CV | 2,58E-05 |
| NLP | 3,60E-06 |

Input and output tokens

To stay consistent with training datasets, we considered the same properties:

**Table 15 - Inputs for Traditional AI tasks.**

| Category | Training dataset considered | Average size (Gb) |
|---|---|---|
| Tabular | Tabular: 1 line * 50 feature * float16 | **1,00E-07** |
| Computer Vision | CV: 1 image * 1920 * 1080 * 3 channels * int8 | **6,22E-03** |
| NLP | NLP: 1 paragraph of 400 tokens | **2,00E-06** |

### GenAI

To estimate the inference usage of LLMs we need to estimate 3 variables:

- The Time to first token (TTFT) (The time in seconds between sending a request to the API and receiving the first token of the response and throughput (The average number of tokens received per second, after the first token is received.) of the models
- The GPUs required to load the model in memory.
- The average size of inputs and outputs

We kept 3 models as reference during our study:

**Table 16 - Reference models for LLMs categories - Generative AI**

| Category | Model used as reference |
|---|---|
| Low (5-15B) | Llama 3.1 8B |
| Medium (40-150B) | Llama 3.1 70B |
| High (+200B or multimodal) | Llama 3.1 405B |

Time to first token (TTFT) / Throughput

We have used the platform Artificial Analysis, which conducts experimental testing of large language models (LLMs) across various API providers. For our analysis, we averaged the TTFT and throughput metrics of the primary cloud providers—Google Cloud Platform (GCP), Amazon Web Services (AWS), and Microsoft Azure—for three reference models.

We have collected both Time-to-First Token (TTFT) and throughput data for input sizes of 100 tokens (used for Chat and Agents) and 1000 tokens (used for RAG).

The speed analysis, summarized in the table below, includes prompt sizes, TTFT, and output speed in tokens per second (tok/s):

**Table 17 - Time to first token and Output speed for Llama 3.1 models for 100 and 1000 tokens prompts**

| Speed tokens analysis | size (B) | Provider | Prompt size tokens | TFFT (s) | Output Speed (tok/s) |
|---|---|---|---|---|---|
| Llama 3.1 8B | 8 | Average (AWS/GCP/Azure) | 100 | 0,26 | 127,0 |
| Llama 3.1 70B | 70 | Average (AWS/GCP/Azure) | 100 | 0,36 | 43,4 |
| Llama 3.1 405B | 405 | Average (AWS/GCP/Azure) | 100 | 0,60 | 21,9 |
| Llama 3.1 8B | 8 | Average (AWS/GCP/Azure) | 1000 | 0,29 | 124,3 |
| Llama 3.1 70B | 70 | Average (AWS/GCP/Azure) | 1000 | 0,43 | 44,0 |
| Llama 3.1 405B | 405 | Average (AWS/GCP/Azure) | 1000 | 0,91 | 21,7 |

## Number of vGPUs used

To estimate the number of vGPUs required to load the model, we have assumed that models are loaded using FP16 precision, with a memory overhead factor of 1.3. The resulting vGPU requirements were rounded up to determine the number of Isolated Multi-Instance NVIDIA A100 GPUs (with up to 10 GB memory each) necessary for deployment.

**Table 18 - Number of Isolated multi-instance GPUs required to load Llama 3.1 models**

| Model Size | FP16 | Nb of vGPU |
|---|---|---|
| Llama 8B | 16 GB | 3 |
| Llama 70B | 140 GB | 19 |
| Llama 405B | 810 GB | 106 |

Note that we did not account for other server emissions (RAM / SSD storage…) as they have been already accounted in our calculation of average vGPU Power consumption. See Life-cycle-assessment methodology for the methodology.

## Input and output tokens

In order to model the input and output sizes of standard use cases such as Chat / RAG and agents. We have gathered various community datasets from which we have extracted the number of tokens from inputs and outputs using Llama 3.1 (8b) tokenizer loaded with AutoTokenizer library.

- **For chat** we have analyzed 4 real-world datasets:

**Table 19 - Average input and output tokens for various chat datasets**

| Dataset | Description | Input tokens mean | Output tokens mean |
|---|---|---|---|
| LM-sys-chat | This dataset contains one million real-world conversations with 25 state-of-the-art LLMs. It is collected from 210K unique IP addresses in the wild on the Vicuna demo and Chatbot Arena website from April to August 2023 | 77,41 | 173,64 |
| OpenAssistant oasst1 | Human-generated assistant-style conversation corpus in 35 different languages. | 30,47 | 178,91 |
| pure_dove | Highly filtered multi-turn conversations between GPT-4 and real humans | 90,69 | 290,00 |
| Share GPT | Filtered version of the ShareGPT dataset, consisting of real conversations between users and ChatGPT. | 309,00 | 191,27 |
| **Average** | | **126,89** | **208,45** |

- **For Agents** we have analyzed Agent-Flan dataset:

**Table 20 - Input and output tokens averages for Agent's datasets**

| Agents | Input length (tokens) | Output length (tokens) | nb of function calls |
|---|---|---|---|
| Agent-Flan | 405,65 | 390,93 | 3,03 |

We also consider that Agents workflows includes calls to external tools, we model these calls by inference to traditional NLP tasks like embedding search, leaving finer modeling for future work.

Note: Input length includes tools description as system prompt. Nb of function calls being the average number of Actions performed by the LLM with external tools.

- **For RAG,** we have based our hypothesis on 3 actual RAG projects in production with our clients. We typically use between 5000 and 6000 tokens for inputs (including system prompt and context), which gives us an average of 5333 input tokens. And results from our projects give 363 averaged output tokens. For RAG we also model the embedding search by 1 supplementary inference of traditional NLP task.

These values are used for storage impact calculation also. We consider that 1 token = 4 characters = 4 bytes.

## Fine tuning

### Traditional AI

<u>Compute</u>

Fine-tuning models for company-specific use cases remains a largely underexplored topic in the literature. Most studies focus on estimating the pre-training resource consumption of base models in NLP[20] or LLMs[26]. Companies usually do not train models from scratch but instead perform fine-tuning starting from pre-trained model checkpoints. Consequently, to model the fine-tuning of AI models for company-specific use cases, we used two different simplified approaches.

- For **Tabular Data**: We have considered the training of a Random Forest Classifier on a dataset of 100k lignes * 50 feature * float16 datapoints. We used Code carbon tracker[23] on a Macbook Pro M2 using 8 CPU for a total of 42.5W and a RAM of 6W. We ran 10 different trainings to get the averaged training consumption for CPUs and RAM. We combined this energy per training with the total number of iterations, training phases in a time-series model life cycle. We estimated the usual number of trainings at 1500 (5 Cross validation per day for 40 days for initial training and one re-training per week during the model usage with a lifetime of 5 years).
- For **NLP/CV**: we have chosen to break down into realistic parameters that would represent the workflow of a Machine learning training. We decided to use figures based on high level hypothesis. We based our estimation on the following parameters:
    - Number of iterations, training phases: The usual number of iterations for a complete training of a model (Estimated, between 10 and 50).
    - Number of epochs during one test (Estimated, between 30 and 70)
    - Batch size during training: (Estimated at 16)
    - Training dataset size:
        - For CV, we considered: 10k images * 1920 * 1080 (size) * 3 rgb channels * int8
        - For NLP: 100 documents * 50 pages * 300 mots * 7 chars which represents ~3750 paragraphs of 400 tokens
    - Energy for 1 inference: we used the same data computed for 1 model inference. See section Inference of supplementary information, note that we multiply by 2 to account for forward and backward calculations.

Note that we have considered the impact at the vCPU and isolated multi-instance vGPU level. A vCPU (Virtual Central Processing Unit) is a virtualized version of a physical CPU core allocated to a virtual machine (VM) in a cloud computing environment. It represents a portion of the physical CPU's resources. Similarly, MIG (Multi-Instance GPU), is a technology developed by NVIDIA that allows a single physical GPU to be partitioned into multiple independent instances, each acting like a smaller, isolated vGPU.

As a result, we get the following figures for total lifetime training energy:

**Table 20 - Energy Training modeling - Traditional AI**

| UC type | Nb of tests | Nb of epochs per test | batch size | Energy compute (Training total, kWh) |
|---|---|---|---|---|
| Tabular | 1500 | 1 | 1 | 2,00E+00 |
| Computer Vision | 15 | 50 | 16 | 2,10E+01 |
| NLP | 20 | 30 | 16 | 8,81E-01 |

<u>Storage</u>

As explained in the last section, we used the representative datasets using usual conversion factors for float/int/char and considering 1 token ~4 characters.

We have also considered a factor 2 to account for postprocessing data.

Finally, we have consider that the data is stored during the whole training process nb of tests * Max(CPU hours per test, GPU hours per test).

**Table 21 - Training datasets - Traditional AI**

| Category | Training dataset considered | Average size (Gb) | Energy storage (Training total, kWh) |
|---|---|---|---|
| Tabular | Tabular: 100k rows * 50 feature * float16 | 2,00E-02 | **1,03E-06** |
| Computer Vision | CV : 10k images * 1920 * 1080 * 3 channels * int8 | 1,24E+02 | **3,22E-03** |
| NLP | NLP : 100 documents * 50 pages * 300 words * 7 chars | 2,10E-02 | **2,34E-08** |

<u>Network</u>

We have considered that the dataset is downloaded once. Resulting in the following energy:

**Table 22 - Energy (kWh) attributed to network usage for Traditional AI use cases training**

| Category | Training dataset considered | Average size (Gb) | Energy network (Training total, kWh) |
|---|---|---|---|
| Tabular | Tabular: 100k rows * 50 feature * float16 | 2,00E-02 | **3,42E-03** |
| Computer Vision | CV : 10k images * 1920 * 1080 * 3 channels * int8 | 1,24E+02 | **2,13E+01** |
| NLP | NLP: 100 documents * 50 pages * 300 words * 7 chars | 2,10E-02 | **3,59E-03** |

Finally, we have applied a lifetime factor of **<u>5 years</u>** for traditional AI models, assuming the models are used consistently over this period. This allows for a more comprehensive estimation of the total energy impact over a year of usage.

### GenAI:

We haven't considered any fine tuning for Generative AI use cases as it still results as a marginal contributor to the overall energy consumption compared to inference tasks. However, as fine-tuning techniques evolve and become more widespread, their impact could become more significant and warrant further analysis in future assessments.

## Company Portfolio Model

### Ratio of AI / Gen AI use cases

Leveraging the work from Nestor Maslej et al.[33] describing the AI vs Generative AI by function, we have computed the ratio of respondents reporting using AI vs GenAI by function and average this result as an estimation of cross function AI vs GenAI adoption.

Using this method, we have found on average that **71%** (6% standard deviation) of deployed use cases are related to AI and **29%** to GenAI.

### Ratio of use case type

For our fictive company portfolio, we have leveraged a list of **350+ AI use cases** developed by Capgemini for various clients across industries. This list of use case was labelled based on project description using a LLM to classify each use case. The proportion of each class is described in Table 23.

**Table 23 - Proportion of AI types use cases in the selected portfolio**

| AI Use case type | Proportion of use cases |
|---|---|
| **Discriminative AI** | **71%** |
| Computer Vision | 11% |
| NLP | 10% |
| Tabular | 79% |
| **Generative AI** | **29%** |
| Agents | 33% |
| Chatbot | 28% |
| RAG | 39% |

### Use Case usage

To the best of our knowledge very few reports address the number of users and usage frequency of an AI use case.

Based on trends in GenAI adoption especially for end users and our experience at Capgemini, we find GenAI use cases usually address a larger audience and are deployed companywide or for specific functions. Users perform weekly to daily requests to the deployed system.

On the other hand, classic AI models are usually tailor made for experts or automated systems. This type of models addresses smaller groups and can either be used very frequently (e.g. for automated systems) or more scarcely on specific expert tasks.

**Table 24 - Distribution of Usage and Usage frequency across clusters**

| Type of AI | Number of users | | | | Usage Frequency | | | |
|---|---|---|---|---|---|---|---|---|
| | Low | Medium | High | Very High | Low | Medium | High | Very High |
| **Classic AI** | 80% | 15% | 5% | 0% | 25% | 25% | 25% | 25% |
| **Gen AI** | 10% | 40% | 30% | 20% | 35% | 40% | 20% | 5% |

## Model Size

As a proxy to estimate the ratio of model size used in a company, we have used the number of downloads on Hugging Face from the last 30 days of the 3 flavors of Llama-3.1 models[18].

However, this does not account for closed source models usage which are supposed to have a larger number of parameters. According to LangChain[62], OpenAI is the most used LLM provider with 6x as much usage as Ollama. Therefore, assuming OpenAI's GPT-4, is 6x more used than Llama's models, we have found the following distribution.

**Table 25 - Distribution of model sizes for generative AI use cases in our 2024 portfolio**

| Model size | Model name | Number of downloads | Ratio of used model size |
|---|---|---|---|
| Low | meta-llama/Llama-3.1-8B-Instruct | 4,788,999 | 13.1% |
| Medium | meta-llama/Llama-3.1-70B-Instruct | 406,486 | 1.1% |
| High | meta-llama/Llama-3.1-405B-Instruct | 30,370 | 85.8% |
| | OpenAI/GPT-4 | ~31,400,000 *(estimated)* | |

## Geographical distribution

To model the distribution of computing resources for a fictitious company, we have used data from a CBRE study[63] on global datacenter inventory in megawatts (MW). The study provided the following capacity distribution:

**Table 26 - Geographical distribution of datacenter power capacity worlwide.**

| Location | Q1 2024 Capacity (MW) | Proportion |
|---|---|---|
| **Americas** | 4900 | *45%* |
| **Europe** | 3050 | *28%* |
| **APAC** | 2140 | *27%* |
| **Total** | **10 090** | |

This distribution was then combined with electricity grid impacts for three geographical regions:

- **Americas**: The majority of compute is located in the United States.
- **Europe**: Datacenter capacities are primarily distributed across London, Frankfurt, Amsterdam, and Paris.
- **APAC**: While specific average grid impacts for the region were unavailable, compute is spread across Japan, Sydney, Singapore, and Hong Kong. For future iterations, incorporating more representative regional data could enhance accuracy.

The electricity grid multi-factor impacts were gathered from Ecoinvent[64] database and presented below.

**Table 27 - Electricity usage multi factors impact per kWh**

| Region | Climate change | Water use | Primary energy use | Resource use |
|---|---|---|---|---|
| | *GWP (kg CO2 eq)* | *WU (m3 eq)* | *PE (MJ)* | *ADPe (kg SB eq)* |
| US | 5,47E-01 | 1,86E-02 | 1,16E+01 | 2,21E-08 |
| CN | 8,71E-01 | 3,82E-02 | 1,56E+01 | 1,12E-08 |
| UE-27 | 4,10E-01 | 1,36E-02 | 1,25E+01 | 2,97E-08 |

## 2030 Systemic projections

In this section we detail the analysis and hypothesis for systemic efficiency inductors.

### PUE

The Power usage effectiveness (PUE) is the amount of power the computing equipment in a data center uses relative to its total energy consumption. According to the latest Uptime institute survey on Datacenters[65] and Google's report of their datacenters' efficiency, the PUE of worldwide datacenters progress limitedly in the past 10 years reaching ~1.1 in 2024 for Google and **1.15 for AWS**[66]. Hence, our baseline scenario with limited efficiency keeps the same PUE at 1.15 and our highly efficient scenario projects a PUE of **1.1** in 2030.

### Grid electricity mix

The Grid electricity mix directly defines the multi-criteria impacts of the operational usage. It depends on the electricity generation means of the countries considered. As a matter of simplification, we have used only CO2 emissions factors variation in our analysis and extrapolating their reduction over the other criteria. Going forward, it is necessary to include all criteria differentiated variations.

**Limited efficiency scenario**: Based on the actual policies of the countries, it is inspired by the World energy outlook of IEA[35].

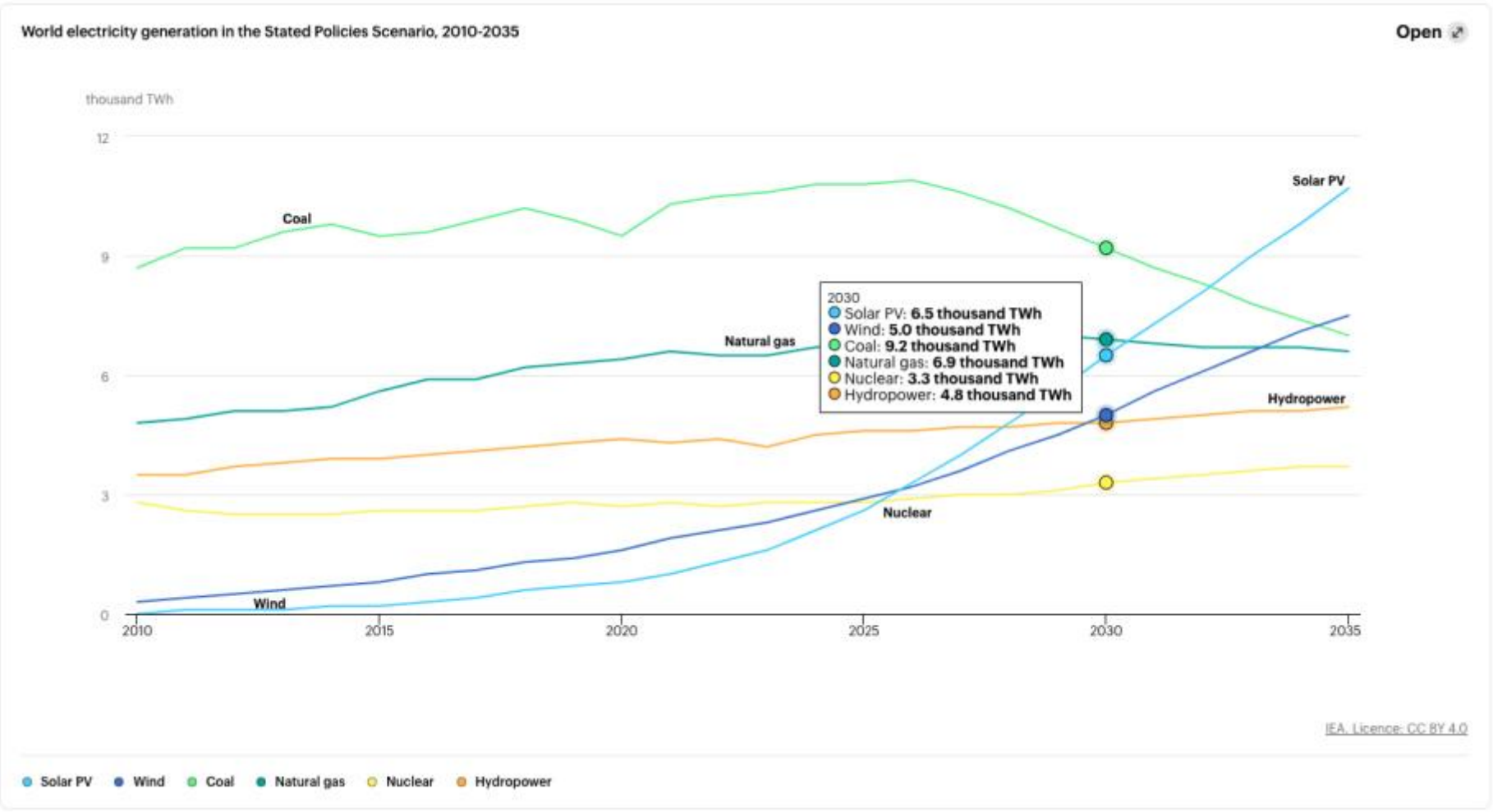

**Figure 6 - World electricity generation in the stated scenario - AIE**

The projected world electricity mix of the AIE projects a -24% of GHG emissions using IPCC 2018 emission factors[67] comparing 2024 and 2030. For simplification, we used the same reduction among the different impacts. On future work, we plan to model differentiated impacts.

**Highly efficient scenario:** Based on IPCC energy systems report[48]: "In scenarios limiting warming to 1.5°C (>50%) with no or limited overshoot (2°C (>67%) with action starting in 2020), net energy system CO2 emissions (interquartile range) fall by 87–97% (60–79%) in 2050. In 2030, in scenarios limiting warming to 1.5°C (>50%) with no or limited overshoot, net CO2 and GHG emissions fall by 35–51% and 38–52% respectively." Consequently, we used efficiency factor of 45% reflecting Paris agreement targets.

Furthermore, additional grid efficiency can be achieved through the independent electricity generation initiatives undertaken by cloud providers, such as the deployment of Small Modular Reactors (SMRs) and large-scale solar farms. By producing their own energy, these providers can reduce their reliance on the public electricity grid, alleviating peak demand pressures and contributing to overall grid stability. Investments are sky rocketing, especially for SMRs.

Predicting the maturity of this technology by 2030 is challenging, as initial deployments are anticipated along this timeline: "Kairos expects to deploy the first SMR by 2030, followed by further deployments through 2035." Consequently, we didn't account for this potential efficiency gains in our scenario.

### Embodied impact

As of now, embodied impacts projections have been left for future directions. Primary ideas stood in modeling the electricity grid impact trajectory of Taiwan as majority of semi-conductors are produced there. Future research should also model the rebound effect related to hardware efficiency gains as producing new components might vary the overall embodied impact. Other studies like Morand and al. [68] have studied the evolution of production impacts of Nvidia workstation graphics cards and would provide insightful projections.

### Hardware efficiency

**Continuous efficiency:** According to epoch AI study[69], hardware is expected to become more power-efficient in 2030. "The peak FLOP/s per W achieved by GPUs used for ML training have increased by around 1.28x/year between 2010 and 2024". If this trend continues linearly, as we can expect from latest Nvidia news about B100 and B300 chips, it results as an increased efficiency x4,4 by 2030.

**Technological breakthrough:** For the high efficiency scenario, we project significant technological breakthroughs that could dramatically accelerate hardware power efficiency by 2030. This outlook anticipates the widespread adoption of next-generation architectures, such as Cerebras inference that runs models like Llama 3.1 20x faster than GPU according to their report[31]. Moreover, we see new appearing technological breakthrough with photonic-powered chips, which hold the potential to revolutionize computational efficiency. Considering these transformative advancements and the increasing adoption of cutting-edge technologies, we project a potential 4.8-fold increase in hardware efficiency by 2030. This estimate is based on a scenario where technologies similar to Cerebras, which are expected to remain 20 times more efficient than GPUs, would account for 20% of total inferences in a high-efficiency scenario.

### Model Efficiency

Three assumptions have driven our model efficiency factors.

Firstly, the growing adoption of **quantization**: "we anticipate more efficient hardware usage in future AI training. While Llama 3.1 405B used FP16 format (16-bit precision), there's growing adoption of FP8 training, as seen with Inflection-2. An Anthropic co-founder has suggested FP8 will become standard practice in frontier labs."[69]. Note that we do not account for power efficiency gains as we cannot predict the GPUs efficiency difference between 16-bit and 8-bit in 2030.

Secondly, the **size of AI models** follows 2 very distinct scenario. This parameter was not modeled in other AI projections studies. For the Limited efficiency scenario, following the race to AGI with bigger and bigger models we assumed that the average size of LLMs models would triple in 2030. From our analysis of lifearchitect.ai data, we calculated that the average size of models doubled from end of 2022 to end of 2024. Moreover, based on the Chinchilla scaling law, which suggests optimal training with one parameter per 20 tokens, models could theoretically reach around 15 trillion parameters by 2030, according to dataset size projections[62]. However, considering recent developments, including speculation around GPT-5 delayed release and increased focus on efficient, reflective architectures, the industry trend may shift towards balancing size and efficiency[71]. Consequently, model size growth may slow down, **tripling by 2030** in our scenario instead of continuous exponential scaling in our limited efficiency scenario. On the other hand, our highly efficient scenario relies on the increasing adoption of Small Language Models and more energy-efficient alternatives[72] alongside the implementation of more frugal practices by companies. Consequently, we assumed a stabilization in model size by 2030, driven by the widespread use of these models.

Thirdly, we modeled the variation of **output tokens** generated by LLMs. Our hypothesis are based on scenario with a <u>1.33</u> increase factor of output tokens for limited efficiency scenario and <u>1.13</u> for more sustainable ones. It reflects the growing interest for reasoning models such as OpenAI o1 and o3 that generate substantially more tokens due to reasoning tokens. [r] showed using various benchmark, that these models generate up to ~1300 tokens for reasoning before generating output that is 5 times more compared with “normal” models. Thus, adoption of these reasoning models for large scale usage may be limited. We assumed for the limited scenario that reasoning models would account for 10% of usage. It results as a similar scaling factor projected as Paccou et al. study[6]. For our usage explosion scenario, we assumed that this rate of adoption would be only possible if AI solutions become way more performant and thus scale in term of complexity. Thus, we project in this high adoption scenario a x3 increase in output size, fueled by growing complexity of chatbots and especially multi-agents frameworks.

**Limited efficiency:** The limited efficiency scenario does not expect any efficiency gain from actual models, assuming FP16 will stay standard, and model size will <u>triple</u> alongside with a growing adoption of reasoning models that will multiply the number of output tokens by <u>1.33</u>.

**High efficiency:** We expect that inference runs will switch over to 8-bit by 2030, which will require <u>2x less</u> numbers of GPUs for loading models. The rise of Small Language models will result in an average size of models stabilization and output tokens size will increase by only 13%.

## 2030 Usage projections

Modeling the adoption dynamics of both Generative and traditional AI presents a complex challenge. To achieve a comprehensive perspective, we have synthesized insights from multiple market analyses conducted by statista[74], Grandview research[75], fortune business, marketsandmarkets[76] along with various report by Capgemini research institute[47] and McKinsey[77]. By integrating these diverse data sources, we have calculated the average Compound Annual Growth Rate (CAGR) used in our Intermediate scenario.

We then outlined two distinct scenarios:

- **Upper Bound Scenario**: A high-adoption trajectory driven by the rapid proliferation of agentic systems, significantly accelerating market penetration.
- **Lower Bound Scenario**: A more conservative expansion, more in line with Data powered enterprises 2024 study by Capgemini Research Institute: "75% of organizations say that large-scale deployment of generative AI PoCs is a significant challenge." [47]

It resulted as the following CAGR from 2024 to 2030:

**Table 28 - Penetration CAGR boundaries from various market analysis.**

| Usage evolution | Low bound | *Associated CAGR* | High bound | *Associated CAGR* |
|---|---|---|---|---|
| **GenAI penetration (excl. Agentic penetration)** | 5,29 | *32 %* | 10,09 | *47 %* |
| **Agentic penetration** | 6,05 | *35 %* | 13,87 | *55 %* |
| **CV AI adoption** | 2,08 | *13 %* | 2,99 | *20 %* |
| **NLP AI Adoption** | 3,30 | *22 %* | 4,83 | *30 %* |
| **Traditional AI adoption** | 2,57 | *17 %* | 3,64 | *24 %* |

# Summary of scenarios

**Table 29 - Boundary scenarios (1/2)**

| Scenario | Factor | Drivers |
|---|---|---|
| **Steady ascent** | **Usage growth** | It reflects the lowest estimations of usage growth for AI found in the various market analysis analyzed. Reflecting more constrained growth highlighted by Grandview research[78].<br>Low boundary usage adoption, resulting in a usage scaling factor of 5.29 for GenAI (exc. Agents), 6,05 for Agents and 2-3 for Traditional AI in 2030. |
| | **Hardware efficiency** | Based on a linear scaling of compute efficiency in FLOPs/W from historical development of Nvidia GPUs. Highlighted in Epoch.AI study[69]. This is our low bound scenario of efficiency, where we do not see any supplementary technological breakthrough except the continuous scaling of GPUs capabilities. It results as efficiency of x1.28 per year according to Epoch.AI and thus x4.4 in 2030. |
| | | PUE is consistent until 2030 at 1.15, in line with previous years trends on PUE stabilization. |
| | **Model Efficiency** | Quantization is not seen a key lever in our limited efficiency scenario. Hence model stay at FP16 precision. |
| | | In our “Do nothing” scenarios, Model providers focus on improving performance and develop bigger and bigger models. Based on our analysis, we project a x3 scaling factor of model sizes in 2030 in these scenarios. |
| | | Similarly, we project a growing complexity of models and thus a growing number of output tokens, driven by the adoption of reasoning models like openAI o1 and o3 resulting in a 33% increase of output sizes. |
| | **Electricity** | Reflecting the World energy outlook 2024 projections on electricity generation of AIE based on countries actual policies.<br>Resulting in a decrease of 24% of GHG emissions of electricity generation. Still far from IPCC targets of -45% in 2030 to achieve 1.5°C limitation. |
| **High adoption without boundaries** | **Usage growth** | This High bound usage adoption scenario is inspired by highest estimations of the market on Generative AI penetration on the market. It reflects market analysis like Statista[79] or Bloomberg Intelligence's projection[80]<br>High boundary usage adoption, resulting in a usage scaling factor of ~10 for GenAI (exc. Agents), 13.9 for Agents and 3-5 for Traditional AI in 2030. |
| | **Hardware efficiency** | Similar to **Steady ascent** |
| | **Model Efficiency** | The primary distinction between the high bound and low bound scenarios lies in how model complexity scales. In this scenario of very high adoption, we anticipate a rapid increase in AI complexity, driving a threefold rise in the volume of output tokens generated by the models. |
| | **Electricity** | Similar to **Steady ascent** |

**Table 30 - Boundary scenarios (2/2)**

| Scenario | Factor | Drivers |
|---|---|---|
| **Limited growth with efficiency breakthrough** | **Usage growth** | Similar to **Steady ascent** |
| | **Hardware efficiency** | For the high efficiency scenario, we project significant technological breakthroughs that could dramatically accelerate hardware power efficiency by 2030. Leveraging on the growing adoption of inference specialized technologies projected at 20% of total inferences in 2030 we project a 4.8-fold increase compared to the "do nothing scenarios". |
| | | PUE slightly decrease of 5% until 2030 at 1.1. |
| | **Model Efficiency** | Quantization is seen a key lever by model providers and hyperscalers. Projecting a 20% usage of int8 quantization for inferences, we achieve an efficiency of x1.2 of efficiency. |
| | | This highly efficient scenario relies on the increasing adoption of Small Language Models and more energy-efficient alternatives alongside the implementation of more frugal practices by companies. Consequently, we assumed a stabilization in model size by 2030, driven by the widespread use of these models. |
| | | Similar to **Steady ascent** |
| | **Electricity** | Reflecting the IPCC targets: "in scenarios limiting warming to 1.5°C (>50%) with no or limited overshoot, net CO2 and GHG emissions fall by 35–51% and 38–52% respectively"<br>Resulting in a decrease of 45% of GHG emissions of electricity generation. |
| **Technological breakthrough** | **Usage growth** | Similar to **High adoption without boundaries** |
| | **Hardware efficiency** | Similar to **Limited growth with efficiency breakthrough** |
| | **Model Efficiency** | We assumed that to fuel the rate of adoption envisaged by this scenario, market will not lie on the same "frugal" practices as the low bound scenario. Hence, model size and complexity scale at the rate than "Do nothing" scenario while quantization is adopted at 20%. |
| | **Electricity** | Similar to **Limited growth with efficiency breakthrough** |

**Table 31 - Intermediate scenario**

| Scenario | Factor | Drivers |
|---|---|---|
| **Intermediate scenario** | **Usage growth** | It reflects the balance of usage adoption between steady growth and High adoption scenarios. It is based on the average of the various markets analysis explored for this study.<br>It results in a usage scaling factor of 7.4 for GenAI (exc. Agents), 9.3 for Agents and 3-5 for Traditional AI in 2030. |
| | **Hardware efficiency** | For this Intermediate scenario, we assume a hardware efficiency increase mainly driven by GPU improvements reflecting Nvidia dominance on the compute market. Efficiency gains scale by 4.4 in 2030 in this scenario. |
| | | PUE is consistent until 2030 at 1.15, in line with previous years trends on PUE stabilization. |
| | **Model Efficiency** | Quantization is not seen a key lever in our limited efficiency scenario. Hence model stay at FP16 precision. |
| | | This Intermediate scenario assumes a balance between highly complex frameworks development and model size increase on one side and collective intelligence in using of GenAI models on the other. Hence, we expect both model size and average output tokens to double in 2030 in this scenario. |
| | **Electricity** | Reflecting the World energy outlook 2024 projections on electricity generation of AIE based on countries actual policies.<br>Resulting in a decrease of 24% of GHG emissions of electricity generation. |

## 2030 scenarios results

**Table 32 - Evolution of number of use cases depending on selected usage scenario**

| Number of use cases | As Is 2024 | Scenario – Usage Increase | Scenario – Explosion of usages | Scenario – Intermediate scenario |
|---|---|---|---|---|
| **Number of use cases** | 100 | 341 | 576 | 486 |
| **Ratio of GenAI use cases** | *29%* | *46%* | *55%* | *46%* |

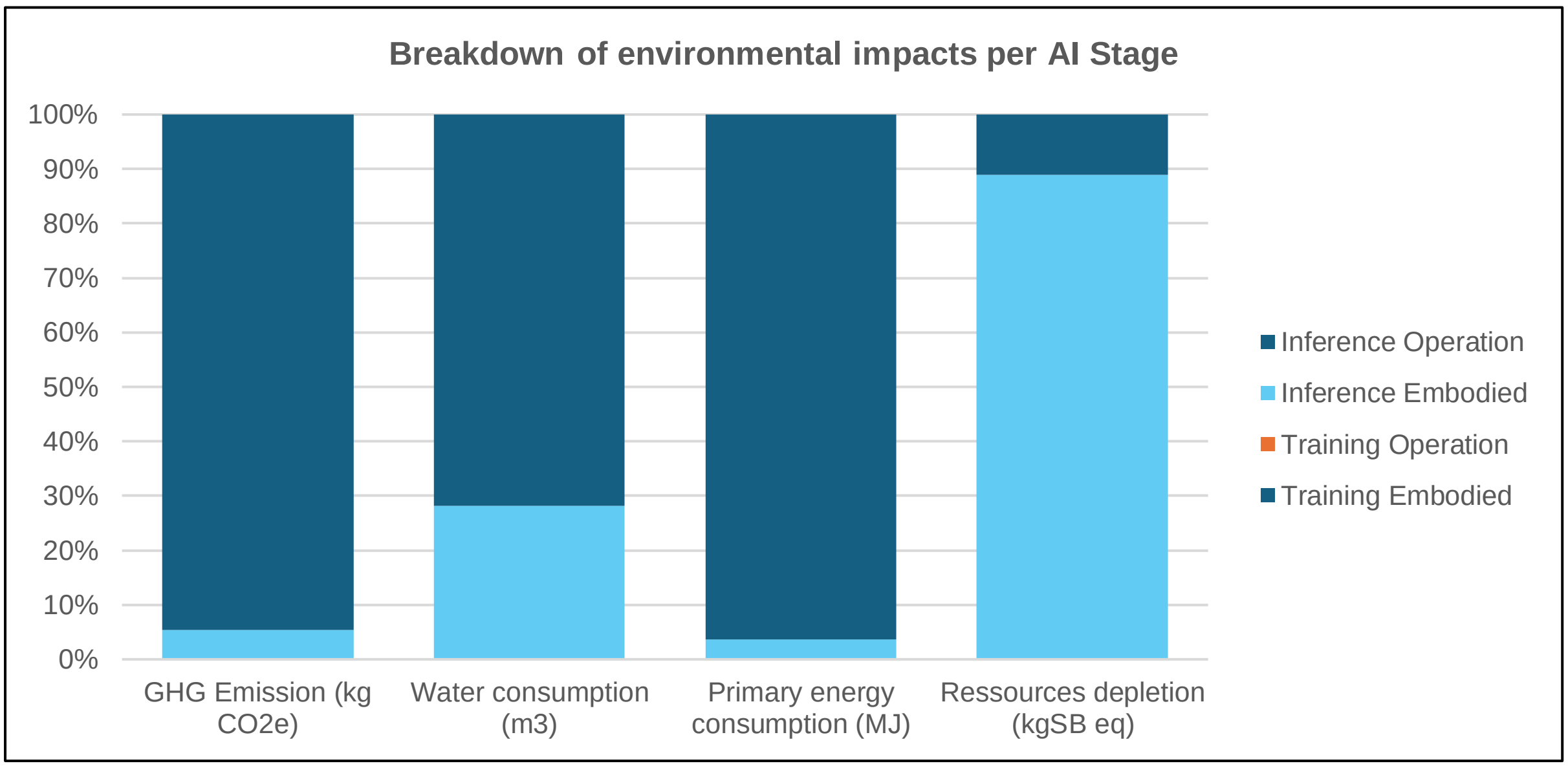

**Figure 7 - Breakdown of environmental impacts per AI Stage from the aggregated 2024 representative portfolio.** It highlights the varied staked of reducing overall impact withing AI footprint. GHG emissions mainly depend on compute efficiency and electricity generation impact while resources depletion is driven by the efficiency of manufacturing compute infrastructure.

## Multi factor impacts summary

**Table 33 - Summary of multi factors impacts of 1 inference.**

| Type of AI | Model size | Use Case type | Operational Climate change (kgCO2eq) | Embodied Climate change (kgCO2eq) | Operational Water use (m3eq) | Embodied Water use (m3eq) |
|---|---|---|---|---|---|---|
| **Gen AI** | Low | Chat | 5,55E-05 | 3,11E-06 | 2,72E-06 | 1,06E-06 |
| **Gen AI** | Medium | Chat | 9,26E-04 | 5,19E-05 | 4,54E-05 | 1,77E-05 |
| **Gen AI** | High | Chat | 1,03E-02 | 5,79E-04 | 5,06E-04 | 1,98E-04 |
| **Gen AI** | Low | RAG | 9,28E-05 | 5,17E-06 | 4,55E-06 | 1,77E-06 |
| **Gen AI** | Medium | RAG | 1,58E-03 | 8,82E-05 | 7,72E-05 | 3,02E-05 |
| **Gen AI** | High | RAG | 1,79E-02 | 1,00E-03 | 8,75E-04 | 3,42E-04 |
| **Gen AI** | Low | Agents | 2,96E-04 | 1,66E-05 | 1,45E-05 | 5,67E-06 |
| **Gen AI** | Medium | Agents | 5,09E-03 | 2,85E-04 | 2,49E-04 | 9,75E-05 |
| **Gen AI** | High | Agents | 5,72E-02 | 3,20E-03 | 2,80E-03 | 1,09E-03 |
| **Trad. AI** | NA | Tabular | 2,06E-08 | 1,37E-09 | 1,01E-09 | 4,40E-10 |
| **Trad. AI** | NA | CV | 1,89E-04 | 8,75E-07 | 9,26E-06 | 2,98E-07 |
| **Trad. AI** | NA | NLP | 2,20E-06 | 1,22E-07 | 1,08E-07 | 4,17E-08 |

| Type of AI | Model size | Use Case type | Operational Primary energy use (MJE) | Embodied Primary energy use (MJ) | Operational Resource use (ADPeq) | Embodied Resource use (ADPeq) |
|---|---|---|---|---|---|---|
| **Gen AI** | Low | Chat | 1,20E-03 | 4,61E-05 | 1,99E-12 | 1,59E-11 |
| **Gen AI** | Medium | Chat | 2,01E-02 | 7,68E-04 | 3,32E-11 | 2,65E-10 |
| **Gen AI** | High | Chat | 2,24E-01 | 8,57E-03 | 3,70E-10 | 2,96E-09 |
| **Gen AI** | Low | RAG | 2,01E-03 | 7,65E-05 | 3,32E-12 | 2,80E-11 |
| **Gen AI** | Medium | RAG | 3,41E-02 | 1,31E-03 | 5,64E-11 | 4,52E-10 |
| **Gen AI** | High | RAG | 3,87E-01 | 1,48E-02 | 6,40E-10 | 5,11E-09 |
| **Gen AI** | Low | Agents | 6,42E-03 | 2,46E-04 | 1,06E-11 | 8,96E-11 |
| **Gen AI** | Medium | Agents | 1,10E-01 | 4,23E-03 | 1,82E-10 | 1,46E-09 |
| **Gen AI** | High | Agents | 1,24E+00 | 4,74E-02 | 2,05E-09 | 1,64E-08 |
| **Trad. AI** | NA | Tabular | 4,47E-07 | 2,05E-08 | 7,39E-16 | 3,17E-13 |
| **Trad. AI** | NA | CV | 4,09E-03 | 1,30E-05 | 6,77E-12 | 1,54E-11 |
| **Trad. AI** | NA | NLP | 4,78E-05 | 1,81E-06 | 7,90E-14 | 2,24E-12 |